\documentclass{article}
\usepackage{ijcai18}

\usepackage{times}
\usepackage[dvipsnames]{xcolor}
\usepackage{soul}
\usepackage[utf8]{inputenc}
\usepackage[small]{caption}
\usepackage{amssymb}
\usepackage{amsmath}

\usepackage{graphicx}

\usepackage{caption}
\usepackage{subcaption}

\usepackage[noend]{algorithm2e}

\newenvironment{tightlist}%
{\begin{list}{$\bullet$}{%
    \setlength{\topsep}{0in}
    \setlength{\partopsep}{0in}
    \setlength{\itemsep}{0in}
    \setlength{\parsep}{0in}
    \setlength{\leftmargin}{1.5em}
    \setlength{\rightmargin}{0in}
    \setlength{\itemindent}{-.1in}
}
}%
{\end{list}
}

\newcommand{\HAC}{{\sc hac}}

\usepackage{hyperref}
\usepackage{multirow}
\usepackage{graphics}
\usepackage{amssymb}
\usepackage{amsthm}

\usepackage{todonotes}

\theoremstyle{definition}
\newtheorem{definition}{Definition}[section]
 
\theoremstyle{remark}

\theoremstyle{plain}
\newtheorem{theorem}{Theorem}[section]
\newtheorem{lemma}[theorem]{Lemma}

\newtheorem{conjecture}[theorem]{Conjecture}
\newtheorem{corollary}{Corollary}[theorem]

\newcommand*{\QEDA}{\hfill\ensuremath{\square}}%

 \newcommand{\Important}[2][1=]{\todo[inline,linecolor=red,backgroundcolor=red!25,bordercolor=red,#1]{#2}}

 \newcommand{\maybe}[2][1=]{\todo[linecolor=orange,backgroundcolor=orange!25,bordercolor=orange,#1]{#2}}
 
 \newcommand{\add}[2][1=]{\todo[linecolor=blue,backgroundcolor=blue!25,bordercolor=blue,#1]{#2}}

\title{Finding Frequent Entities in Continuous Data}

\author{
Ferran Alet, 
Rohan Chitnis, 
Leslie P. Kaelbling, 
Tom\'{a}s Lozano-P\'{e}rez
\\ 
MIT Computer Science and Artificial Intelligence Laboratory \\
\{alet, ronuchit, lpk, tlp\}@mit.edu
}

\begin{document}

\maketitle
\begin{abstract}
In many applications that involve processing high-dimensional data, it is important to identify a small set of entities that account for a significant fraction of detections. Rather than formalize this as a clustering problem, in which all detections must be grouped into hard or soft categories, we formalize it as an instance of the \textit{frequent items} or \textit{heavy hitters} problem, which finds groups of tightly clustered objects that have a high density in the feature space. We show that the \textit{heavy hitters} formulation generates solutions that are more accurate and effective than the clustering formulation. In addition, we present a novel online algorithm for heavy hitters, called \HAC, which addresses problems in continuous space, and demonstrate its effectiveness on real video and household domains.

\end{abstract}


\section{Introduction}

Many applications require finding entities in raw data, such as individual objects or people in image streams or particular speakers in audio streams. Often, entity-finding tasks are addressed by applying clustering algorithms such as $k$-means (for instance in \cite{Niebles2008}).  We argue that instead they should be approached as instances of the {\em frequent items problem}, also known as the {\em heavy hitters problem}.  The classic frequent items problem assumes discrete data and involves finding the most frequently occurring items in a stream of data. We propose to generalize it to continuous data.

\begin{figure}[t]
    \centering
    \begin{subfigure}[t]{\linewidth}
    \begin{center}
        \includegraphics[width=\linewidth]{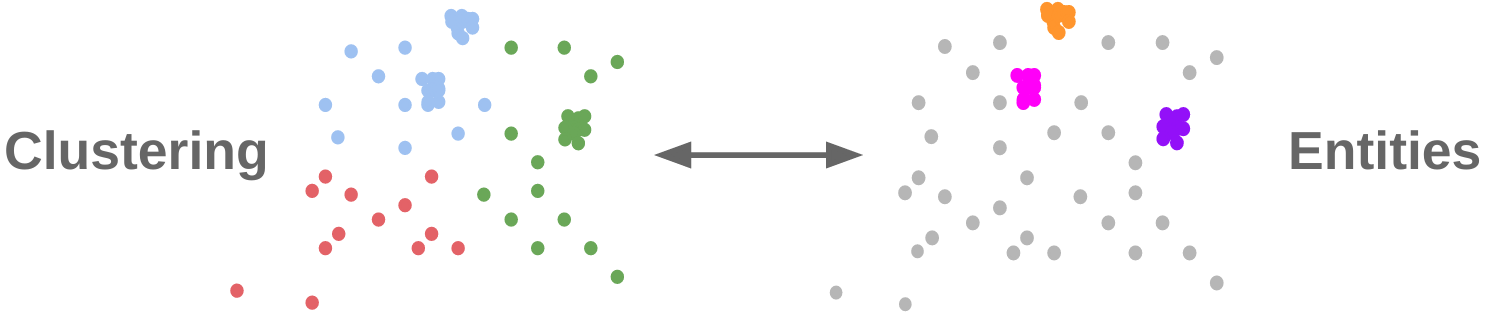}
    \end{center}
    \caption[width=\linewidth]{\textbf{Noisy environment}: outliers (red points) greatly influence clustering. Entities, defined locally, are robust to large amounts of noise.}
    \label{fig:noisy_env}
    \end{subfigure}
    \begin{subfigure}[t]{\linewidth}
    \begin{center}
    \centering
    \includegraphics[width=\linewidth]{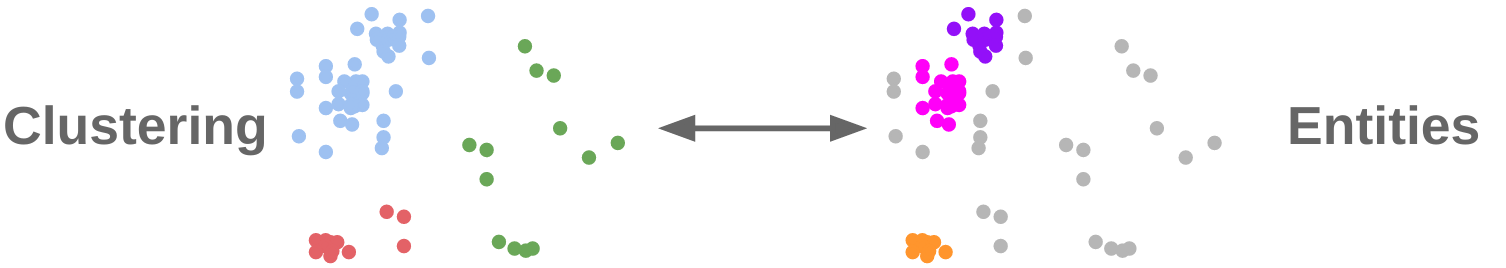}    
    \end{center}
    \caption[width=\linewidth]{\textbf{Groups of different sizes}: clustering tends to join nearby groups; entities may be close together yet distinct. }
    \label{fig:join_groups}
    \end{subfigure}
    \vspace{-.5\baselineskip}
    \caption{In clustering \textit{all/most} points belong to a group, forming \textit{big} clusters which are defined \textit{globally}. In entity finding \textit{some} points belong to a group, forming \textit{small tight} regions defined \textit{locally}.}
    \label{fig:clust_vs_entities}
    \vspace{-1\baselineskip}
\end{figure}

Figure \ref{fig:clust_vs_entities} shows examples of the differences between clustering and entity finding. Some clustering algorithms fit a global objective assigning \textit{all/most} points to centers, whereas entities are defined \textit{locally} leading to more robustness to noise (\ref{fig:noisy_env}). Others, join nearby dense groups while trying to detect sparse groups, whereas entities are still distinct (\ref{fig:join_groups}). These scenarios are common because real world data is often noisy and group sizes are often very unbalanced \cite{newman2005power}.

We characterize entities using two natural properties: \textbf{similarity} - the feature vectors should be similar according to some (not necessarily Euclidean) distance measure, such as cosine distance, and \textbf{salience} - the region should include a sufficient number of detections over time.

Even though our problem is not well-formulated as a clustering problem, it might be tempting to apply clustering algorithms to it. Clustering algorithms optimize for a related, but different, objective. This makes them less accurate for our problem; moreover, our formulation overcomes typical limitations of some clustering algorithms such as relying on the Euclidean distance metric and performing poorly in high-dimensional spaces. This is important because many natural embeddings, specially those coming from Neural Networks, are in high dimensions and use non-Euclidean metrics. 

In this paper we suggest addressing the problem of entity finding as an extension of heavy hitters, instead of clustering, and propose an algorithm called \HAC\ with multiple desirable properties: handles an online stream of data; is guaranteed to place output points near high-density regions in feature space; is guaranteed to \emph{not} place output points near low-density regions (i.e., is robust to noise); works with \textit{any} distance metric; can be time-scaled, weighting recent points more; is easy to implement; and is easily parallelizable.

We begin by outlining a real-world application of tracking important objects in a household setting without any labeled data and discussing related work. We go on to describe the algorithm and its formal guarantees and describe experiments that find the main characters in video of a TV show and that address the household object-finding problem.

\subsection{Household Setting}
The availability of low-cost, network-connected cameras provides an opportunity to improve the quality of life for people with special needs, such as the elderly or the blind. One application is helping people to find misplaced objects.

More concretely, consider a set of cameras recording video streams from some scene, such as a room, an apartment or a shop. At any time, the system may be queried with an image or a word representing an object, and it has to answer with candidate positions for that object. Typical queries might be: \textit{"Where are my keys?"} or \textit{"Warn me if I leave without my phone."} Note that, in general, the system won't know the query until it is asked and thus cannot know which objects in the scene it has to track. For such an application, it is important for the system to not need specialized training for every new object that might be the focus of a query.  

Our premise is that images of interesting objects are such that 1) a
neural network embedding~\cite{Donahue,Johnson,Mikolov} will place them close together in feature space, and 2) their position stays constant most of the time, but changes occasionally. Therefore objects will form
high-density regions in a combined feature$\times$position space.  Random
noise, such as people moving or false positive object detections, will not
form dense regions.  Objects that don't move (walls, sofas, etc) will
be always dense; interesting objects create dense regions in
feature$\times$position space, but eventually change position and form a new
dense region somewhere else.  We will exploit the fact that our algorithm
is easy to scale in time, to detect theses changes over time. 

\subsection{Related work}
Our algorithm, \HAC, addresses the natural generalization of \textit{heavy hitters}, a very well-studied problem, to continuous settings. In heavy hitters we receive a stream of elements from a discrete vocabulary and our goal is to estimate the most frequently occurring elements using a small amount of memory, which does not grow with the size of the input. Optimal algorithms have been found for several classes of heavy hitters, which are a logarithmic factor faster than our algorithm, but they are all restricted to \textit{discrete} elements \cite{manku2002approximate}. In our use case (embeddings of real-valued data), elements are not drawn from a discrete set, and thus repetitions have to be defined using regions and distance metrics. Another line of work \cite{Chena} estimates the total number of different elements in the data, in contrast to \HAC\ that finds (not merely counts) different \textit{dense} regions.  

Our problem bears some similarity to clustering but the problems are fundamentally different (see figure \ref{fig:clust_vs_entities}). The closest work to ours within the clustering literature is  \textit{density-based (DB) clustering}. In particular, they first find all dense regions in space (as we do) and then join points via paths in those dense regions to find arbitrarily-shaped clusters. In contrast, we only care about whether a point belongs to one of the dense regions. This simplification has two advantages: first, it prevents joining two close-by entities, second, it allows much more efficient, general and simple methods.

The literature on DB clustering is very extensive. Most of the popular algorithms, such as DBScan \cite{Ester} and Level Set Tree Clustering \cite{Chaudhuri}, as well as more recent algorithms \cite{Rodriguez2014}, require simultaneous access to all the points and have complexity quadratic in the number of points; this makes them impractical for big datasets and specially streaming data. There are some online DB clustering algorithms \cite{Chen}, \cite{Wan2009},\cite{Cao}, but they either tessellate the space or assume a small timescale,  tending to work poorly for non-Euclidean metrics and high dimensions.

Two pieces of work join ideas from clustering with \textit{heavy hitters}, albeit in very different settings and with different goals. \cite{larsen2016heavy} uses graph partitioning to attack the discrete $l_p$ \textit{heavy hitters} problem in the general turnstile model. \cite{braverman2017clustering} query a \textit{heavy hitter} algorithm in a tessellation of a high dimensional discrete space, to find a coreset which allows them to compute an approximate $k$-medians algorithm in polynomial time. Both papers tackle streams with discrete elements and either use clustering as an intermediate step to compute \textit{heavy hitters} or use \textit{heavy hitters} as an intermediate step to do clustering ($k$-medians). In contrast, we make a connection pointing out that the generalization of \textit{heavy hitters} to continuous spaces allows us to do \textit{entity finding}, previously seen as a clustering problem.
\begin{figure}[t]
    \centering
    \begin{subfigure}[t]{0.22\textwidth}
    \begin{center}
        \includegraphics[width=\linewidth]{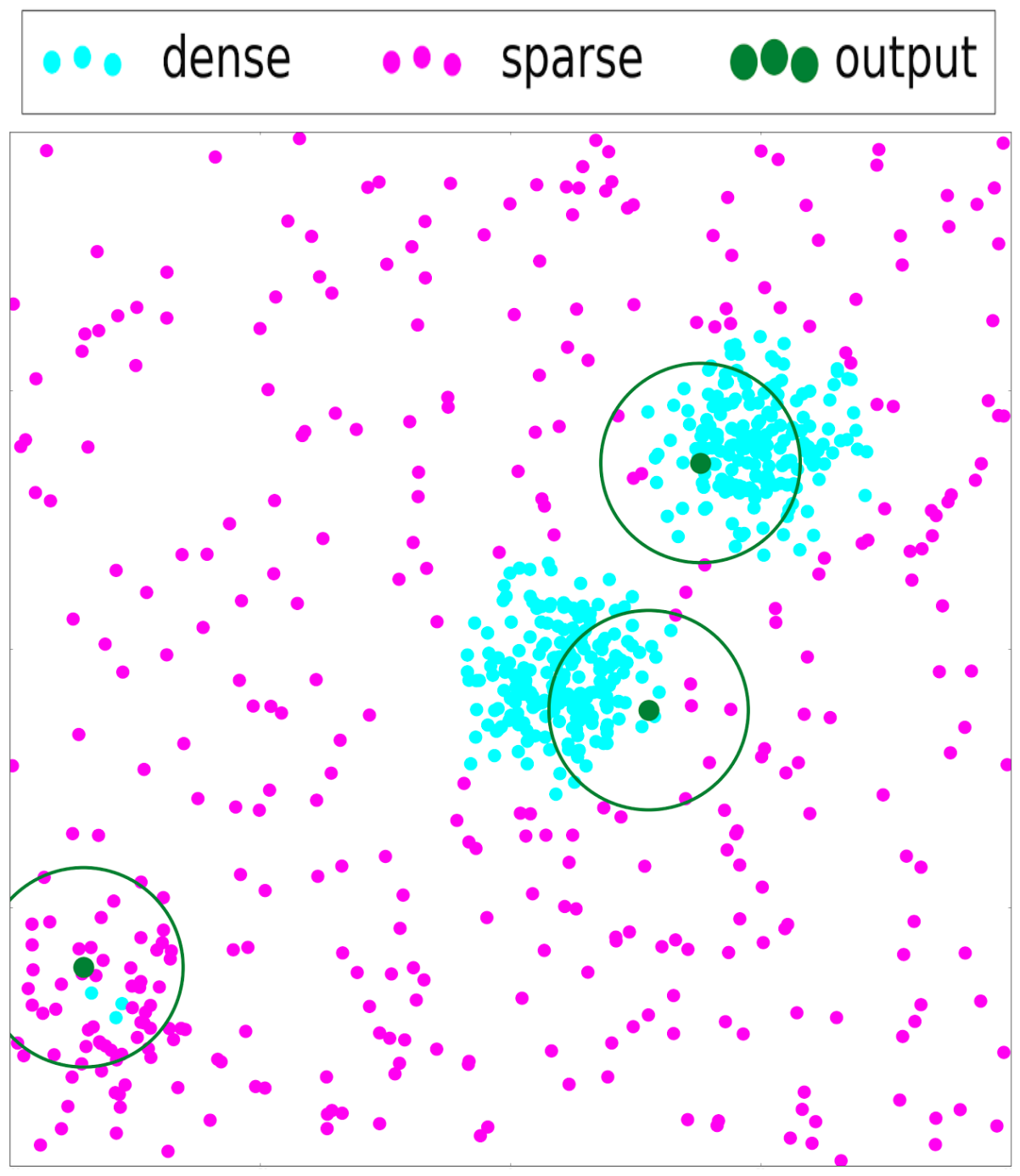}
    \end{center}
    \caption[width=\linewidth]{$f=7\%$}
    \label{fig:rf_definition_low_f}
    \end{subfigure}
    \hfill
    \begin{subfigure}[t]{0.22\textwidth}
    \begin{center}
    \centering
    \includegraphics[width=\linewidth]{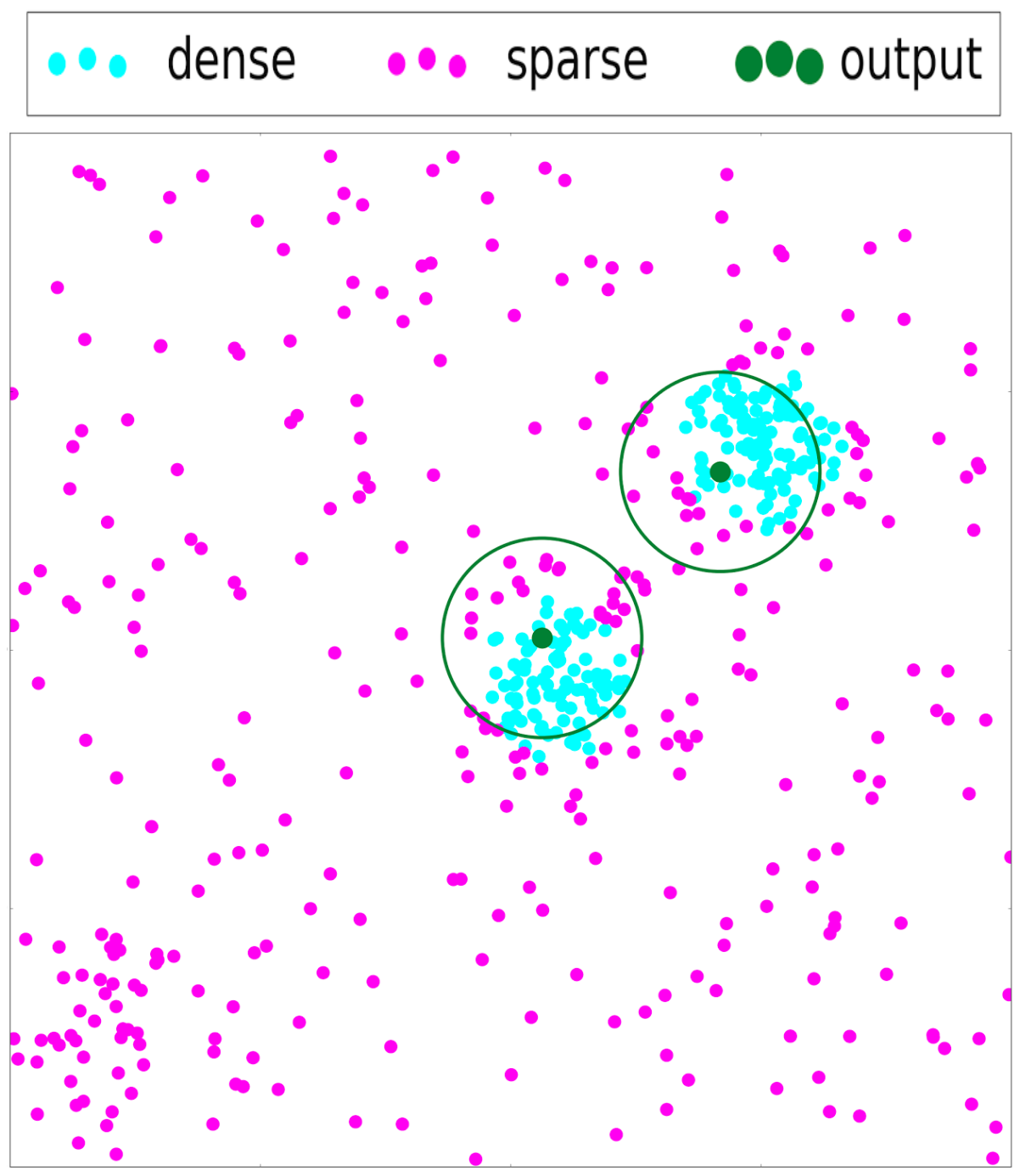}    
    \end{center}
    \caption[width=\linewidth]{$f=15\%$}
    \label{fig:rf_definition_high_f}
    \end{subfigure}
    
    \caption[Intuition of $(r,f)$ dense for varying fraction $f$]{\textbf{Varying fraction $f$} with fixed radius $r$. Data comes from 3 Gaussians plus uniform random noise. A circle of radius $r$ near the sparsest Gaussian captures more than $7\%$ of the data but less than $15\%$; thus being dense in (a), but not in (b).}
    \label{fig:clust_vs_entities}
    
    \includegraphics[width=.45\linewidth]{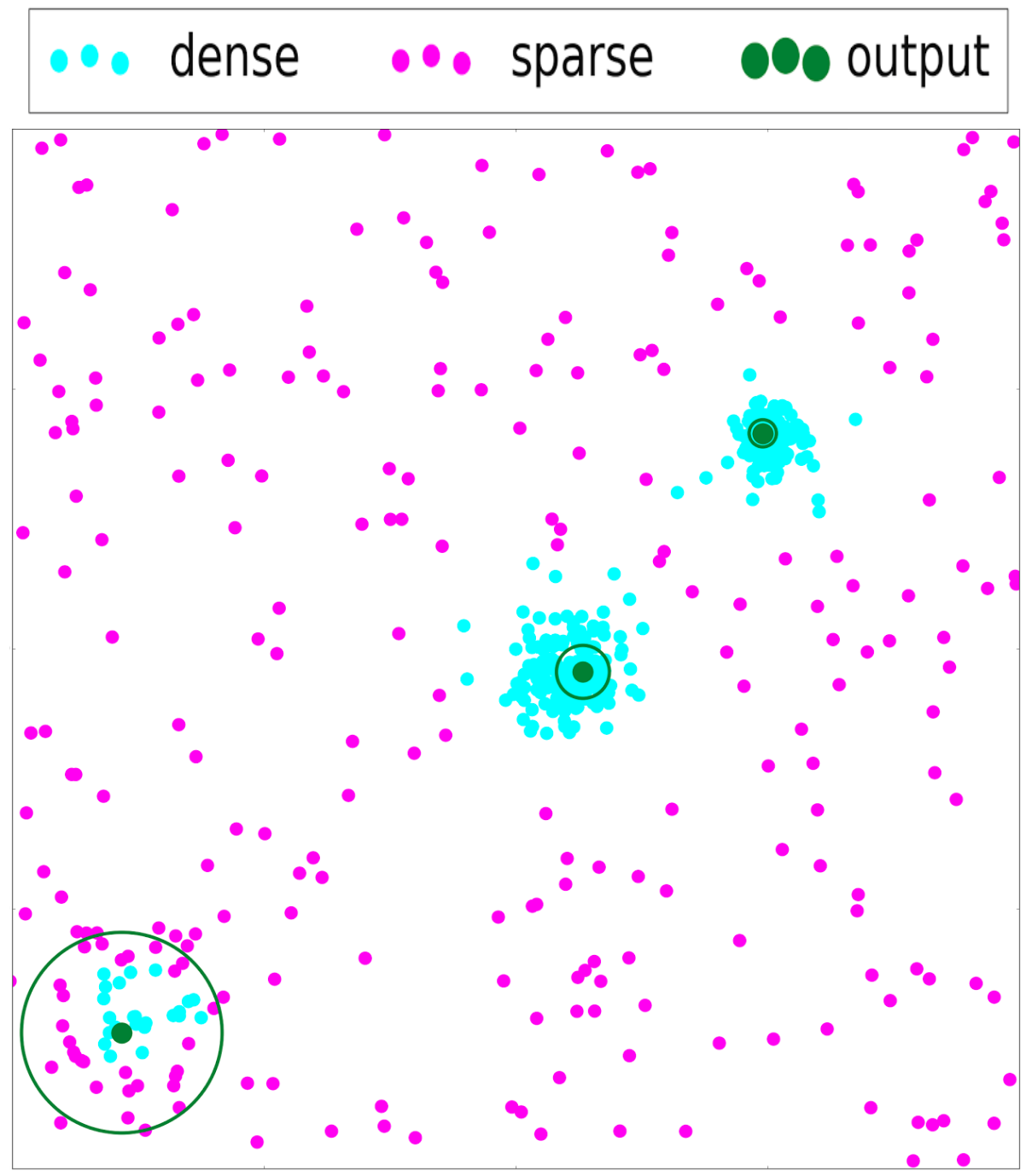}
    \caption[Intuition of $(r,f)$ dense for varying radius $r$]{\textbf{Varying radius $r$} with fixed frequency $f$. We can detect Gaussians with different variances by customizing $r$ for each output. \\The goal \textit{isn't} to cover the whole group with the circle but to return the smallest radius that contains a fraction $f$ of the data. Points near an output are guaranteed to need a similar radius to contain the same fraction of data.}
    \label{fig:clust_vs_entities}
    \vspace{-1\baselineskip}
\end{figure}


We illustrate our algorithm in some applications that have been addressed using different methods.  Clustering faces is a well-studied problem with commercially deployed solutions. However, these applications generally assume we care about most faces in the dataset and that faces
occur in natural positions. This is not the case for many real-world applications, where photos are taken in motion from multiple angles and
are often blurry. Therefore, algorithms that use clustering in the conventional sense, \cite{Schroff,Otto2017}, do not apply. 

\cite{Rituerto2016} proposed using \textit{DB}-clustering in a setting similar to our object localization application. However, since our algorithm is online, we allow objects to change position over time. Their method, which uses DBScan, can be used to detect what we will call \textit{stable objects}, but not movable ones (which are generally what we want to find). 
\cite{Nirjon} built a system that tracks objects assuming they will only change position when interacting with a human. However, they need an object database, which makes the problem easier and the system much less practical, as the human has to register every object to be tracked. 
\newcommand\mycommfont[1]{\smaller\ttfamily\textcolor{blue}{#1}}
\SetCommentSty{mycommfont}

\section{Problem setting}

In this section we argue that random sampling is surprisingly effective (both theoretically and experimentally) at finding entities by detecting \textit{dense} regions in space and describe an algorithm for doing so in an online way. The following definitions are of critical importance.

\begin{definition}
Let  $d(\cdot, \cdot)$ be the distance metric. A point $p$ is \textit{$(r,f)$-dense} with respect to dataset $\mathcal{D}$ if the subset of points in $\mathcal{D}$ within distance $r$ of $p$ represents a fraction of the points that is at least $f$. If $N = |\mathcal{D}|$; then $p$ must satisfy:
$$|\{x\in \mathcal{D} \mid d(x,p) \leq r\}| \geq f N.$$
\end{definition}
\begin{definition}
A point $p$ is \textit{$(r,f)$-sparse} with respect to dataset $\mathcal{D}$ if and only if it is not $(r,f)$-\textit{dense}.
\end{definition}

The basic version of our problem is the natural generalization of \textit{heavy hitters} to continuous spaces. Given a metric $d$, a frequency threshold $f$, a radius $r$ and a stream of points $\mathcal{D}$, after each input point the output is a set of points. Every $(r,f)$-\textit{dense} point (even those not in the dataset) has to be close to at least one output point and every $(r,f/2)$-\textit{sparse} region has to be far away from all output points.

Our algorithm is based on samples that \textit{hop} between data points and \textit{count} points nearby; we therefore call it \textit{Hop And Count (\HAC)}.  

\subsection{Description of the algorithm}
A very simple \textit{non}-online algorithm to detect dense regions is to take a random sample of $m$ elements and output only those samples that satisfy the definition of \textit{$(r,f)$-dense} with respect to the whole data set. For a large enough $m$, each dense region in the data will contain at least one of the samples with high probability, so the output will include a sample from this region. For sparse regions, even if they contain a sampled point, this sample will not be in the output since it will not pass the denseness test.

Let us try to make this into an online algorithm. A known way to maintain a
uniform distribution in an online fashion is \textit{reservoir
  sampling}\cite{vitter1985random}: we keep $m$ \textit{stored samples}. After the $i$-th point
arrives, each sample changes, independently, with probability $1/i$ to this new
point. At each time step, samples are uniformly distributed over
all the points in the data. However, once a sample has been drawn we
cannot go back and check whether it belongs to a dense or sparse region
of space, since we have not kept all points in memory. 

The solution is to keep a counter for each sample in memory and update
the counters every time a new point arrives. In particular, for any
sample $x$ in memory, when a new point $p$ arrives we check whether
$d(x,p)\leq r$; if so, we increase $x$'s
counter by 1. When the sample hops to a new point $x'$, the counter is
no longer meaningful and we set it to 0. 

Since we are in the online setting, every sample only sees points that
arrived after it and thus only the first point in a region sees all
the other points in that region. Therefore, if we want to detect a
region containing a fraction $f$ of the data, we have to introduce an acceptance threshold lower than $f$, for example $f/2$, and only output points with frequency above it. The probability of any sample being in the first half of any
dense region is at least $f/2$ and thus, for a large enough number of
samples $m$, with high probability every dense region will contain a
sample detected as dense.  Moreover, since we set
the acceptance threshold to $f/2$, regions much sparser than $f$ will not
produce any output points. In other words, we will have false
positives but they will be \textit{good} false positives, since those
points are guaranteed to be in regions almost as dense as the target
dense regions we actually care about. In general we can change $f/2$ to $(1-\epsilon)f$ with $\epsilon$ trading memory with performance. Finally, note that this algorithm is easy to parallelize because all samples and their counters are completely independent.


\subsection{Multiple radii}
In the previous section we assumed a specific known threshold $r$. What if we don't know $r$, or if every dense region has a different
diameter?  We can simply have counts for multiple values of $r$ for
every sample. In particular, for every $x$ in memory we maintain a count of streamed points within distance $r$
for every $r\in \{r_0=r_{\min}, r_0\gamma, r_0\gamma^2, \dots,
r_0\gamma^c = r_{\max}\}$. At output time we can output the smallest $r_i$ such that the $x$ is $(r_i, f)$-\textit{dense}. With this exponential sequence we guarantee
a constant-factor error while only losing a logarithmic factor in memory usage. $r_0$ and $c$ may be user-specified or automatically adjusted at runtime.

Following is the pseudo-code version of the algorithm with multiple specified radii.  Note that the only data-dependent parameters are $r_0$ and $c$, which specify the minimum and maximum radii, and $f_0$ which specifies the minimum fraction that we will be able to query.  The other parameters ($\epsilon$, $\delta$, $\gamma$) trade off memory vs. probability of statisfying guarantees.
\begin{algorithm}[h]
\begin{small}
\SetAlgoLined
\DontPrintSemicolon
  \SetKwFunction{algo}{algo}\SetKwFunction{proc}{proc}
  \SetKwProg{myalg}{Algorithm}{}{}
  \SetKwProg{myproc}{Subroutine}{}{}
  \myalg{Hop And Count Processing($f_0$, $\epsilon$, $\delta$, $r_0$, $\gamma$, $c$)}{
  \nl $m \leftarrow \log{(f_0^{-1}\delta^{-1}})/f_0\epsilon$ \tcp*[f]{to satisfy guarantees} \; 
  \nl $Mem \leftarrow [\emptyset, \overset{(m)}{\dots}, \emptyset]$ ; $Counts \leftarrow Zeros(m, c)$\;
  \nl $t = 0$ \;
  \nl \For{$p$ in stream} {
    \nl $t\ +=1$\;
    \nl \For{$0\leq i\leq m$} {
    \nl \If{$Bernoulli(1/t)$} {
    \nl $Mem[i] \leftarrow p$ \tcp*[f]{hop}\;
    \nl \For{$0\leq r \leq c$} {
    \nl $Counts[i][r] \leftarrow 0$ \tcp*[f]{reset counters}\;}}
    \nl $r \leftarrow \textbf{max}\left(0,\textbf{ceil}\left(\log_\gamma\left(d(Mem[i],p)/r_0\right)\right)\right)$\;
    \nl \If{$r \leq c$}{
    \nl $Counts[i][r]\ +=1$\;}
    }
  }
  }
\myalg{Hop And Count Query($f$, $t$, $\epsilon$, $Mem$, $Counts$)}{
    \nl \For(\tcp*[f]{$0\leq i < m$}){$0\leq i <$ len(Counts) } {
        \nl $count \leftarrow 0$ \;
        \For(\tcp*[f]{$0\leq r < c$}){$0\leq r < len(Counts[i])$}{
            \nl $count \leftarrow count + Mem[i][r]$ \;
            \nl \If{$count\geq (1-\epsilon)ft$}{
                \nl \textbf{output} $\left(Mem[i], r\right)$\;
                \nl \textbf{break}\;
            }
        }
    }
  }
\end{small}
    \vspace{-\baselineskip}
\end{algorithm}

\subsection{Guarantees}
\label{subsec:guarantees}
We make a guarantee for every dense or sparse point in space, even
those that are not in the dataset. Our guarantees are probabilistic;
they hold with probability $1-\delta$ where $\delta$ is a parameter of
the algorithm that affects the memory usage. We have three types of guarantees, from loose but very
certain, to tighter but less certain. 
 For simplicity, we assume here that
$r_{\min}=r_{\max}=r$.   Here, we state the theorems;
the proofs are available in appendix \ref{appendix:sec:appendix}.

\begin{definition}
$r_f(p)$ is the smallest $r$ s.t. $p$ is $(r,f)$-dense. For each point $p$ we refer to its \textit{circle/ball} as the sphere of radius $r_f(p)$ centered at $p$.
\end{definition}

\begin{theorem}
    For any tuple $(\epsilon<1, \delta, f)$, with probability $1-\delta$, for any point $p$ s.t. $r_f\leq r_{\max}/2\gamma$ our algorithm will give an output point $o$ s.t. $d(o,p)\leq 3r_f(p)$.\\
Moreover, the algorithm always needs at most $\Theta\left(\frac{\log (f\delta)}{\epsilon f}\log_\gamma\left(\frac{r_{\max}}{r_{\min}}\right)\right)$  memory and $\Theta(\frac{\log (f\delta)}{\epsilon f})$ time per point. Finally, it outputs at most $\Theta\left(\frac{\log (f\delta)}{\epsilon f}\right)$ points.
\label{theorem:whp}
\end{theorem}

\begin{lemma}
Any $(\Delta,(1-\epsilon)f)$-sparse point will not have an output point within $\Delta-2r_{\max}$.
\label{lemma:triangular}
\end{lemma}

Notice that we can use this algorithm as a noise detector with provable guarantees. Any $(r_{\max}, f)$-\textit{dense} point will be within $3r_{\max}$ of an output point and any $(5r_{\max}, (1-\epsilon)f)$-\textit{sparse} point will not.

\begin{theorem}
    For any tuple $(\epsilon, \delta, f)$, with probability $(1-\delta)$, for any $(r,f)$-dense point $p$ our algorithm will output a point $o$ s.t. $d(o,p) \leq r$ with probability at least $(1-\delta f)$.
\label{theorem:most-points}
\end{theorem}

\begin{theorem}
\label{thm:awesome-postprocess}
We can apply a post-processing algorithm that takes parameter $\gamma$ in time $\Theta\left(\frac{\log (f\delta)}{\epsilon f^2}\right)$ to reduce the number of output points to $(1+2\epsilon)/f$ while guaranteeing that for any point $p$ there is an output within $(4\gamma + 3)r_f(p)$. 
The same algorithm guarantees that for any $(r_{max}, f)$-\textit{dense} point there will be an output within $7r_{max}$.
\end{theorem}
Note that the number of outputs can be arbitrarily close to the optimal $1/f$.

The post-processing algorithm is very simple: iterate through the original outputs in increasing $r_f(p)$. Add $p$ to the final list of outputs $O$ if there is no $o\in O$ s.t. $d(o,p)\leq r_f(p)+r_f(o)$. See appendix \ref{appendix:sec:appendix} for a proof of correctness.

In high dimensions many clustering algorithms fail; in contrast, our performance can be shown to be provably good in high dimensions. We prove asymptotically good performance for dimension $d\rightarrow \infty$ with a convergence fast enough to be meaningful in real applications.

\begin{theorem} With certain technical assumptions on the data distribution, if we run \HAC\ in high dimension $d$, for any $(r,1.05f)$-dense point there will be an output point within $(1+\alpha)r$, with $\alpha=O(d^{-1/2})$, with probability $(0.95-\delta f-O(e^{-fn}))$, where $n$ is the total number of datapoints.\\
Moreover, the probability that a point $p$ is $(r, 0.98(1-\epsilon)f)$-\textit{sparse} yet has an output nearby is at most $0.05+O(e^{-fn})$. 
\label{thm:soft-high-dim-inf-dist}
\end{theorem}
We refer the reader to appendix \ref{appendix:sec:appendix} for a more detailed definition of the theorem and its proof.

The intuition behind the proof is the following: let us model the dataset as a set of high-dimensional Gaussians plus uniform noise. It is well-known that most points drawn from a high dimensional Gaussian lie in a thin spherical shell. This implies that all points drawn from the same Gaussian will be similarly dense (have a similar $r_f(p)$) and will either all be \textit{dense} or all \textit{sparse}. Therefore, if a point is $(r,f)$-\textit{dense} it is likely that another point from the same Gaussian will be an output and will have a similar radius. Conversely, a point that is $(r,(1-\epsilon)f)$-\textit{sparse} likely belongs to a \textit{sparse} Gaussian and no point in that Gaussian can be detected as \textit{dense}.

Note that, for $d,n\rightarrow \infty $ and $\delta\rightarrow 0$ the theorem guarantees that any $(r,f)-$\textit{dense} point will have an output within $r$ with probability $95\%$ and any $(r,(1-\epsilon))$-\textit{sparse} point will not, with probability $5\%$; close to the ideal guarantees. Furthermore, in the appendix we show how these guarantees are non-vacuous for values as small as $n=5000, d=128$: the values of the dataset in section \ref{sec:people-id}.

\definecolor{good-color}{RGB}{106,168,79}
\definecolor{wrong-color}{RGB}{224,102,102}
\definecolor{duplicate-color}{RGB}{230,159,0}
\definecolor{old-duplicate-color}{RGB}{241,194,50}
\definecolor{blue-duplicate-color}{RGB}{109,158,235}
\definecolor{missing-color}{RGB}{103,78,167}

\begin{figure*}[t]
    \centering
    \includegraphics[width=.99\linewidth]{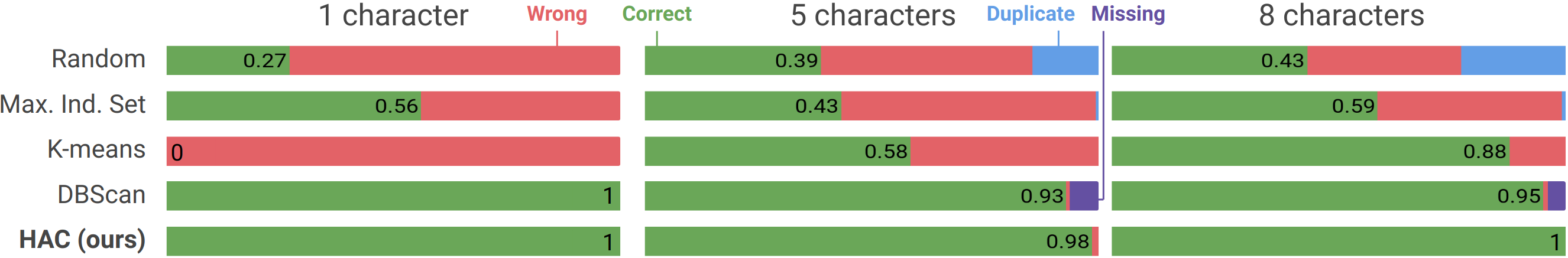}
    \caption{\textbf{Identifying the main $n$ characters for $n\in\{1,5,8\}$.} We ask each algorithm to give $n$ outputs and compute the fraction of main $n$ characters found in those $n$ outputs. We report the average of 25 different random seeds sampling the original dataset for $70\%$ of the data. There are 3 ways of missing: {\color{wrong-color} \textbf{Wrong}}: a noisy image (such as figure \ref{fig:k-means1}) or unpopular character, {\color{blue-duplicate-color} \textbf{Duplicate}}: an extra copy of a popular character, {\color{missing-color} \textbf{Missing}}: the algorithm is unable to generate enough outputs. Despite being online, \HAC\ outperforms all baselines.
    }
    \label{fig:results_tv}
    \vspace{-\baselineskip}
\end{figure*}
\subsection{Time scaling}
We have described a time-independent version of \HAC\ in which all points have equal weight, regardless of when they arrive. However, it is simple and useful to extend this algorithm to make point $i$ have weight proportional to $e^{-(t-t_i)/\tau}$ for any timescale $\tau$, where $t$ is the current time and $t_i$ is the time when point $i$ was inserted.

Trivially, a point inserted right now will still have weight $1$. Now, let $t'$ be the time of the last inserted point. We can update all the weights of the previously received points by a factor $e^{-(t-t')/\tau}$. Since all the weights are multiplied by the same factor, sums of weights can also be updated by multiplying by $e^{-(t-t')/\tau}$.

We now only need to worry about hops. We can keep a counter for the total weight of the points received until now. Let us define $w_{j, k}$ as the weight of point $p_j$ at the time point $k$ arrives. Since we want to have a uniform distribution over those weights, when the $i$-th point arrives we simply assign the probability of hopping to be $1/\sum_{j\leq i} w_{j, i}$. Note that for the previous case of all weights being $1$ (i.e. $\tau = \infty$) this reduces to a probability of $1/i$ as before.

We prove in the appendix that by updating the weights and modifying the hopping probability, the time-scaled version has guarantees similar to the original ones.

\subsection{Fixing the number of outputs}
We currently have two ways of querying the system: 1) Fix a single distance $r$ and a frequency threshold $f$, and get back all regions that are $(r,f)$-dense; 2) Fix a frequency $f$, and return a set of points $\{p_i\}$, each with a different radius $\{r_i\}$ s.t. a point $p$ near output point $p_i$ is guaranteed to have $r_f(p) \approx r_i$.

It is sometimes more convenient to directly fix the number of outputs instead. With \HAC\ we go one step further and return a list of outputs sorted according to density (so, if you want $o$ outputs, you pick the first $o$ elements from the output list). Here are two ways of doing this: 1) Fix radius $r$. Find a set of outputs $p_i$ each $(r,f_i)$-dense. Sort $\{p_i\}$ by decreasing $f_i$, thus returning the densest regions first. 
2) Fix frequency $f$, sort the list of regions from smallest to biggest $r$. Note, however, that the algorithm is given a fixed memory size which governs the size of the possible outputs and the frequency guarantees.

In general, it is useful to apply duplicate removal. In our experiments we sort all $(r,f)$-\textit{dense} outputs by decreasing $f$, and add a point to the final list of outputs if it is not within $r_d$ of any previous point on the list. This is similar to but not exactly the same as the method in theorem \ref{thm:awesome-postprocess}; guarantees for this version can be proved in a similar way.

\def\checkmark{\tikz\fill[scale=0.4](0,.35) -- (.25,0) -- (1,.7) -- (.25,.15) -- cycle;} 
\section{Identifying people}
\label{sec:people-id}

\begin{figure}[t]
    \centering
    \begin{subfigure}[t]{0.17\textwidth}
    \includegraphics[width=\linewidth]{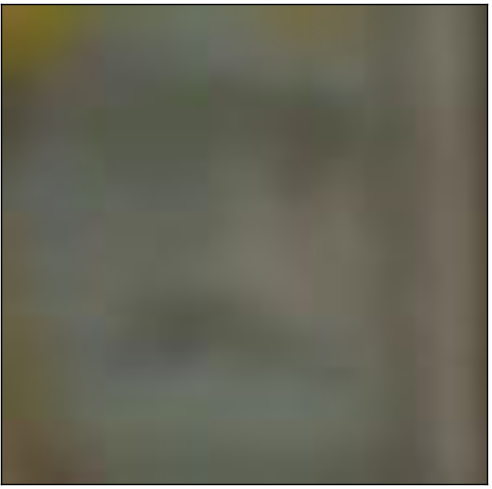}
    \caption{}
    \label{fig:k-means1}
    \end{subfigure}
    \begin{subfigure}[t]{0.28\textwidth}
    \includegraphics[width=1\linewidth]{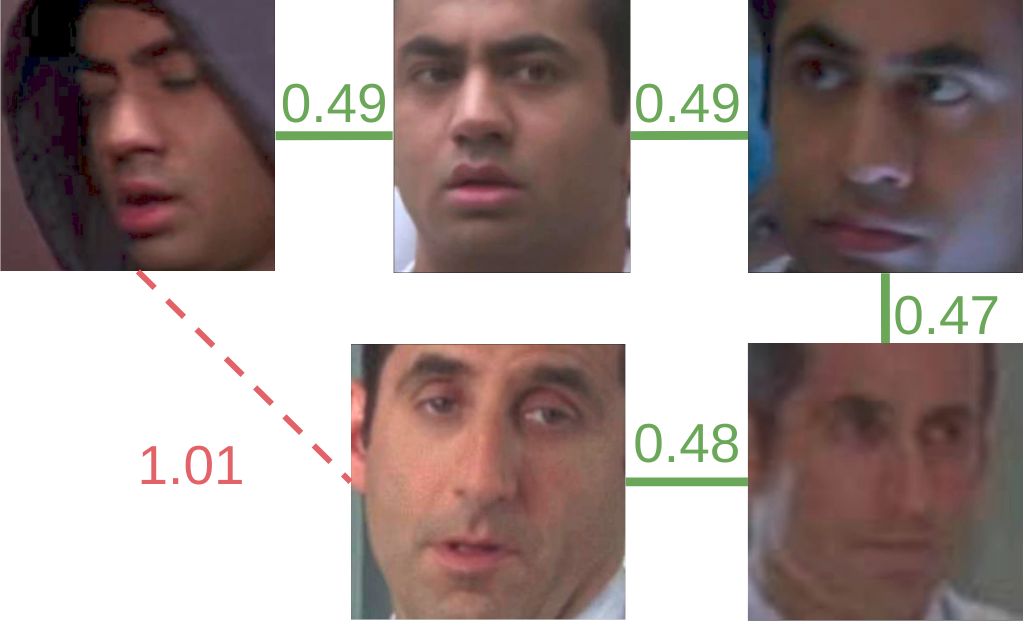}
    \caption{}
    \label{fig:path_dbscan}
    \end{subfigure}
    \vspace{-.5\baselineskip}
    \caption{Shortcomings of clustering algorithms in \textit{entity finding}.\\ (a) The closest training example to the mean of the dataset (1-output of $k$-means) is a blurry misdetection. (b) \textit{DBSCAN} merges different characters through paths of similar faces.}
    \vspace{-\baselineskip}
    \label{fig:14_centers}
\end{figure}

As a test of \HAC's ability to find a few key entities in a large, noisy dataset, we analyze a season of the TV series {\em House M.D.}. We pick 1 frame per second and run a face-detection algorithm (dlib \cite{dlib09}) that finds faces in images and embeds them in a 128-dimensional space. Manually inspecting the dataset reveals a main character in $27\%$ of the images, a main cast of four characters appearing in $6\%$ each and three secondary characters in $4\%$ each. Other characters account for $22\%$ and poor detections (such as figure \ref{fig:k-means1}) for $25\%$.

We run \HAC\ with $r=0.5$ and apply duplicate reduction with $r_d=0.65$. These parameters were \textit{not} fine-tuned; they were picked based on comments from the paper that created the CNN and on figure \ref{fig:guarantees}.
We fix $\epsilon=\delta=0.5$ for all experiments; these large values are sufficient because \HAC\ works better in high dimensions than guaranteed by theorem \ref{theorem:whp}.
\definecolor{good-color}{RGB}{106,168,79}
\begin{figure}[h]
    \centering
    \includegraphics[width=.7\linewidth]{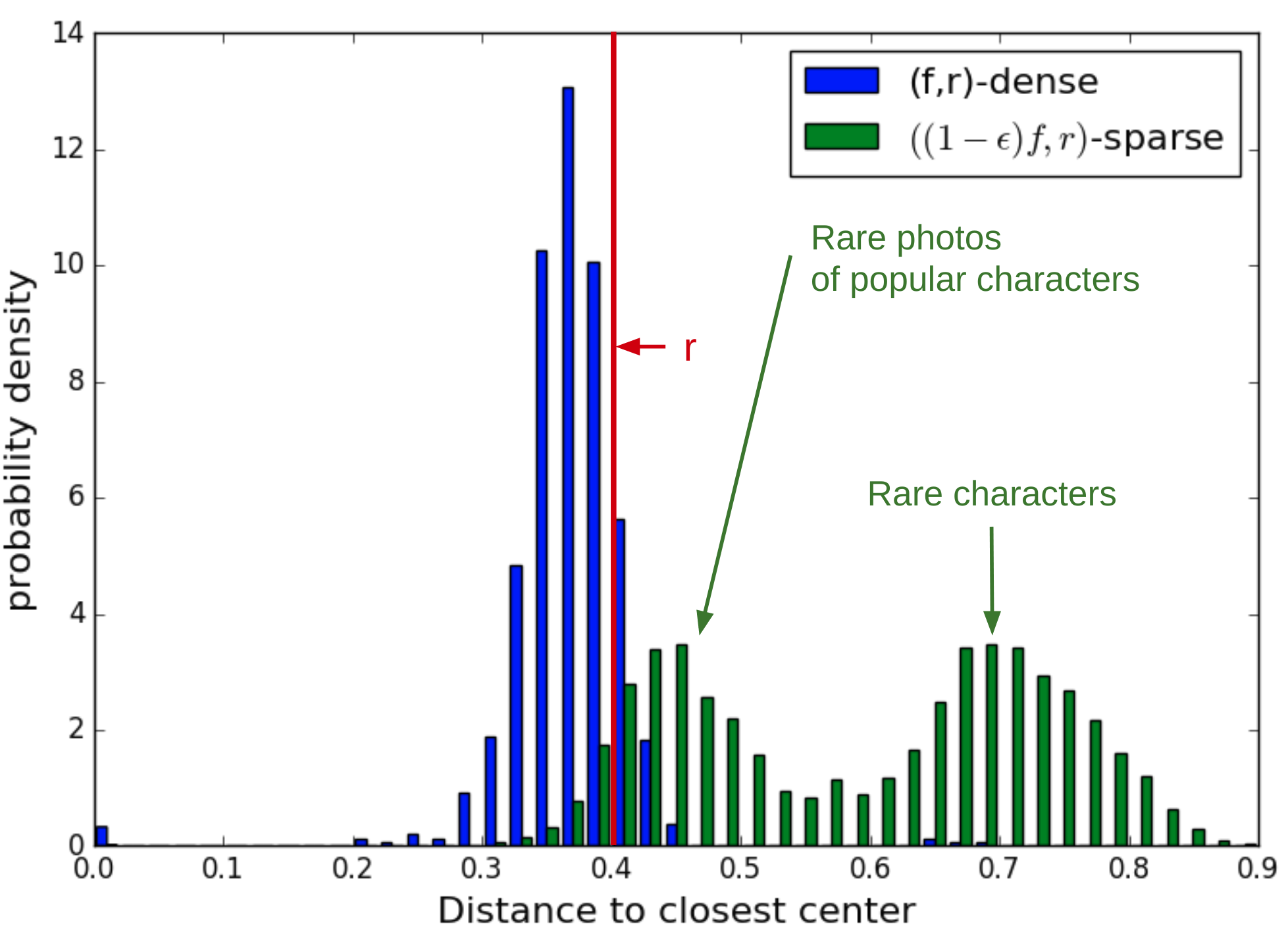}
    \caption{\textbf{Most $(r,f)$-\textit{dense} points are within $r$ of an output, most $(r,(1-\epsilon)f)$-\textit{sparse} points are not,} as predicted by theorem \ref{thm:soft-high-dim-inf-dist}.\\ We run \HAC\ with f=0.02, $r=0.4$. We compare two probability distributions: distance to the closest output for {\color{blue} \textit{dense}} points and for {\color{good-color}  \textit{sparse}} points. Ideally, we would want all the dense points (blue distribution) to be to the left of the threshold $r$ and all the \textit{sparse} points (green) to be to its right; which is almost the case. \\Moreover, notice the two peaks in the frequency distribution (intra-entity and inter-entity) with most uncertainty between 0.5 and 0.65. }
    \label{fig:guarantees}
    \vspace{-1\baselineskip}
\end{figure}
We compare \HAC\ against several baselines to find the most frequently occurring characters. For $n=\{1,5,8\}$ we ask each algorithm to return $n$ outputs and check how many of the top $n$ characters it returned. The simplest baseline, \textit{Random}, returns a random sample of the data. \textit{Maximal Independent Set} starts with an empty list and iteratively picks a random point and adds it to the set iff it is at least $r=0.65$ apart from all points in the list. We use \textit{sklearn} \cite{scikit-learn} for both $k$-means and \textit{DBSCAN}. \textit{DBSCAN} has two parameters: we set its parameter $r$ to $0.5$, since its role is exactly the same as our $r$ and grid-search to find the best $\epsilon$. For $k$-means we return the image whose embedding is closer to each center and for \textit{DBSCAN} we return a random image in each cluster. 

As seen in figure \ref{fig:results_tv}, \HAC\ consistently outperforms all baselines. In particular, $k$-means suffers from trying to account for most of the data, putting centers near unpopular characters or noisy images such as figure \ref{fig:k-means1}. \textit{DBSCAN}'s problem is more subtle: to detect secondary characters, the threshold frequency for being dense needs to be lowered to $4\%$. However, this creates a path of \textit{dense regions} between two main characters, joining the two clusters (figure \ref{fig:path_dbscan}). 

While we used \textit{offline} baselines with fine-tuned parameters, \HAC\ is \textit{online} and its parameters do not need to be fine-tuned. Succeeding even when put at a disadvantage, gives strong evidence that \HAC\ is a better approach for the problem.

Finally, with this data we checked the guarantees of theorem \ref{thm:soft-high-dim-inf-dist}: most $(f,r)$-dense points have an
output within distance $r$, $95\%$, whereas few $(r,(1-\epsilon))$-sparse points do: $6\%$. This is shown in figure \ref{fig:guarantees}.

\section{Object localization}
\label{sec:obj-local}
\label{sec:objects}
In this section we show an application of \textit{entity finding} that cannot be easily achieved using clustering. We will need the flexibility of \HAC: working online, with arbitrary metrics and in a time-scaled setting as old observations become irrelevant.
\subsection{Identifying objects}

In the introduction we outlined an approach to object localization that does not require prior knowledge of which objects will be queried. To achieve this we exploit many of the characteristics of the \HAC\ algorithm.  We assume that: 1) A convolutional neural network embedding will place images of the same object close together and images of different objects far from each other. 2) Objects only change position when a human picks them up and places them somewhere else.

\begin{figure}
    \centering
    \includegraphics[width=.9\linewidth]{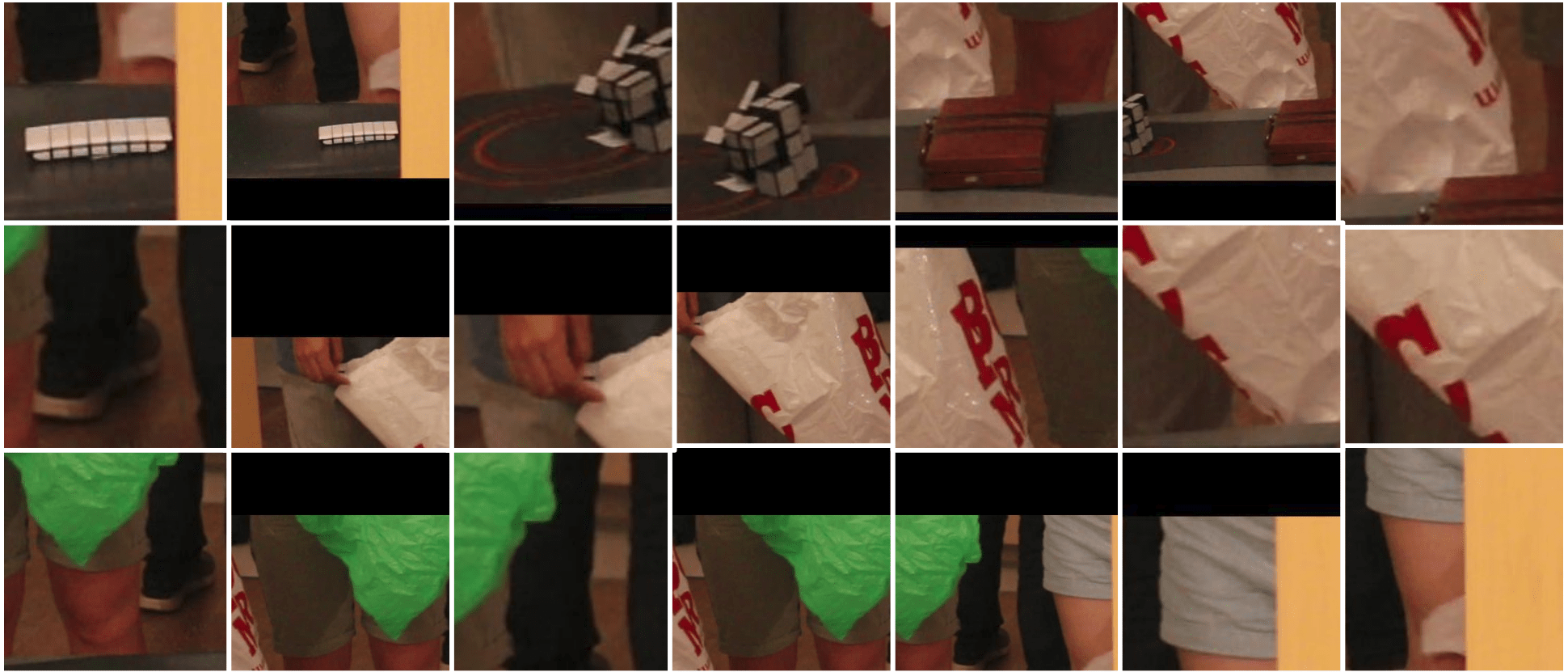} 
    \caption{All the candidate objects from a random camera and time. Only a few proposals (first 6) capture objects of actual interest.}
    \label{fig:sample_detections}
    \vspace{-1\baselineskip}
\end{figure}

Points in the data stream are derived from images as follows.  First, we use SharpMask\cite{Pinheiro2016} to segment the image into patches containing object candidates (figure \ref{fig:sample_detections}). Since SharpMask is not trained on our objects, proposals are both unlabeled and very noisy. For every patch, we feed the RGB image
into a CNN (Inception-V3 \cite{Szegedy}),
obtaining a 2048-dimensional embedding. We then have 3
coordinates for the position (one indicates which camera is used, and
then 2 indicate the pixel in that image).

We need a distance for this representation.  It is natural to assume that two patches represent the same object if their embedding features are similar \textit{and} they are close in the 3-D world. We can implement this with a metric that is the maximum between the distance in feature space and the distance in position space:
$$d((p_1, f_1), (p_2, f_2)) = \max(d_{p}(p_1, p_2),d_{f}(f_1,f_2))$$
We can use cosine distance for $d_f$ and $l_2$ for $d_p$; \HAC\ allows for the use of arbitrary metrics.  However, for good performance, we need to scale the distances such that close in position space and close in feature space correspond to roughly similar numerical values.

We can now apply \HAC\ to the resulting stream of points.  In contrast to our previous experiment, time is now very important. In particular, if we run \HAC\ with a large timescale $\tau_l$ and a small timescale $\tau_s$, we'll have 3 types of detections:
\begin{tightlist}
    \item Noisy detections (humans passing through, false positive camera detections): not dense in either timescale;
    \item Detections from stable objects (sofas, walls, floor): dense
      in both timescales; and 
    \item Detections from objects that move intermittently (keys,
      mugs): not dense in $\tau_l$, and alternating dense and
      sparse in $\tau_s$. (When a human picks up an object from a
      location, that region will become sparse; when the human
      places it somewhere else, a new region will become dense.) 
\end{tightlist}

{\setlength{\parindent}{0pt} We are mainly interested in the third type of detections.}

\subsection{Experiment: relating objects to humans}
We created a dataset of 8 humans moving objects around 20 different locations in a room; you can find it on \url{http://lis.csail.mit.edu/alet/entities.html}. Locations were spread across 4 tables with 8, 4, 4, 4 on each respectively. Each subject had a bag and followed a script with the following pattern: Move to the table of location A; Pick up the object in your location and put it in your bag; Move to the table of location B; Place the object in your bag at your current location.

The experiment was run in steps of 20 seconds: in the first 10 seconds humans performed actions, and in the last 10 seconds we recorded the scene without any
actions happening. Since we're following a script and humans have finished their actions, during the latter 10 seconds we know the position of every object
with an accuracy of 10 centimeters. The total recording lasted for 10 minutes and each human picked or placed an object an average of 12 times. In front of
every table we used a cell phone camera to record that table (both human faces and objects on the table). 

We can issue queries to the system such as: Which human has touched each object? Which objects have not been touched? Where can I find a particular object?
Note that if the query had to be answered based on only the current camera image, two major issues would arise: 1) We would not know whether an object is relevant to a human. 
2) We would not detect objects that are currently occluded.
\begin{figure}
    \centering
    \includegraphics[width=\linewidth]{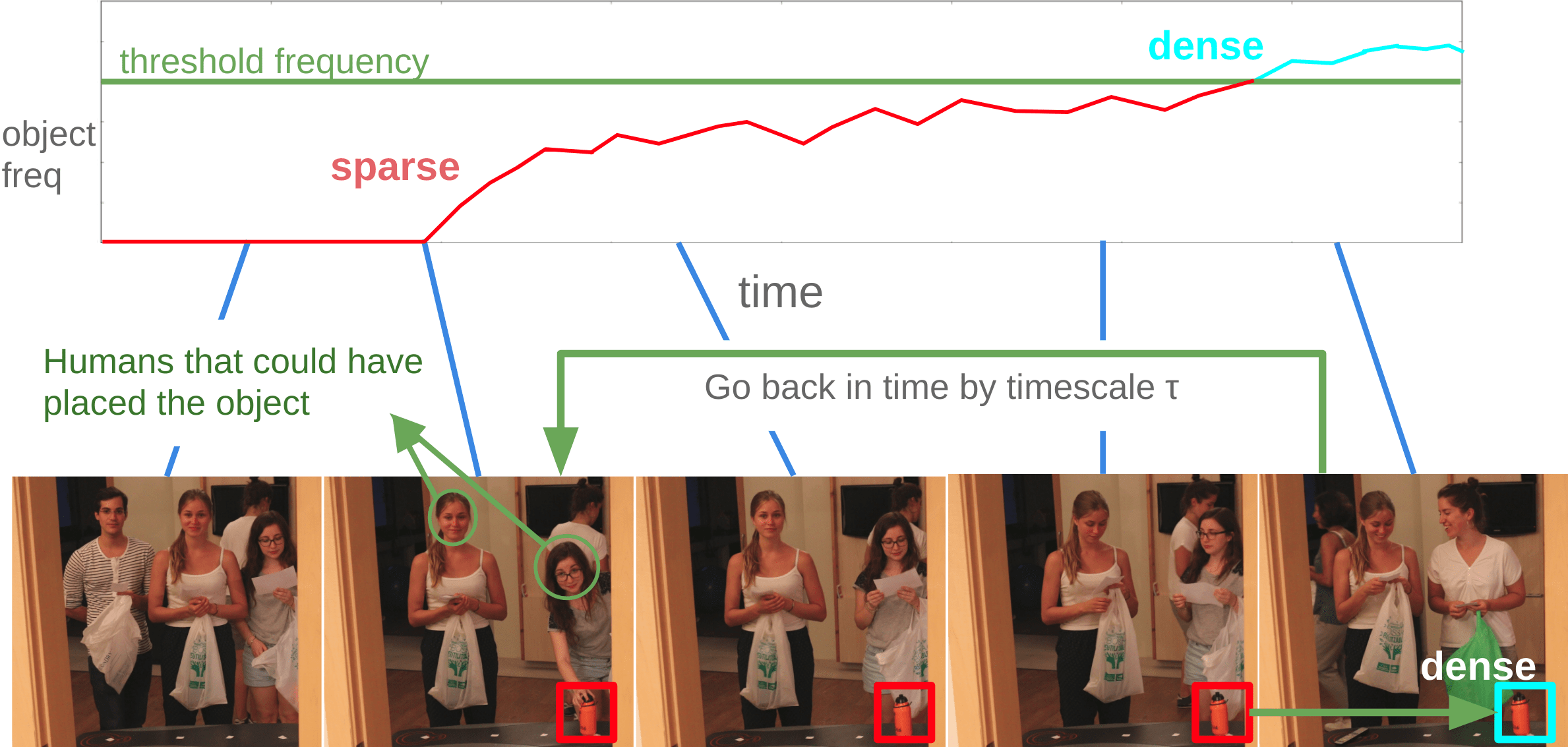}
    \caption{When an object is placed, its frequency starts growing. It takes on the order of the timescale $\tau$ to reach its stationary value, surpassing the threshold frequency. When an object becomes dense/sparse we assume a human placed/picked it, go $\tau$ back time and mark the pair $(obj,human)$. This system is completely unlabeled; $obj$ and $human$ are both just feature vectors.}
    \label{fig:humanscouldhaveplaced}
    \vspace{-1\baselineskip}
\end{figure}

This experimental domain is quite challenging for several reasons: 1) The face detector only detects about half the faces. Moreover, false negatives are very correlated, sometimes missing a human for tens of seconds. 2) Two of the 8 subjects are identical twins. We have checked that the face detector can barely tell them apart. 3) The scenes are very cluttered: when an interaction happens, an average of 1.7 other people are present at the same table. 4) Cameras are 2D (no depth map) and the object proposals are very noisy.

We focus on answering the following query: for a given object, which human interacted with it the most? The algorithm doesn't know the queries in advance nor is it provided training data for particular objects or humans. Our approach, shown in figure \ref{fig:humanscouldhaveplaced}, is as follows:

\begin{tightlist}
    \item Run \HAC\ with $\tau_l = \infty$ (all points have the same weight regardless of their time), $\tau_s = 10$ seconds, $f=2.5\%$ and a distance function and threshold which link two detections that happen roughly within 30 centimeters and have features that are close in embedding space.
    \item Every $10s$, query for outputs representing dense regions.
    \item For every step, look at all outputs from the algorithm and check which ones do not have any other outputs nearby in the previous step. Those are the detections that \textit{appeared}. Similarly, look at the outputs from the previous step that do not have an output nearby in the current step; those are the ones that \textit{disappeared}.
    \item (Figure \ref{fig:humanscouldhaveplaced}) For any output point becoming dense/sparse on a given
      camera, we take its feature vector (and drop the position); call these features $v$ and the current time
      $t$. We then retrieve all detected faces for that camera at
      times $[t-2\tau_s, t-\tau_s]$, which is when a human should have
      either picked or placed the object that made the dense region
      appear/disappear. For any  face $f_i$ we add the pair
      $(v,f_i)$ to a list with a score of $1/|f_i|$, which
      aims at distributing the responsibility of the action between
      the humans present. 
\end{tightlist}
Now, at query time we want to know how much each human interacted with each object. We pick a representative picture of every object and every human to use as queries. 
We compute the pair of feature vectors, compare against each object-face pair in the list of interactions and sum its weight if both the objects and the faces are close. This estimates the number of interactions between human and object.

Results are shown in table~\ref{tab:res_objects}.  There is one row per object.  For each object, there was a true primary human who interacted with it the most.  The columns correspond to:  the number of times the top human interacted with the object, the number of times the system predicted the top human interacted with the object, the rank of the true top human in the predictions, and explanations. \HAC\ successfully solves all but the extremely noisy cases, despite being a hard dataset and receiving no labels and no specific training.

\begin{footnotesize}
\begin{table}[]
    \centering
    \begin{tabular}{c|c|c|l}
         \#pick/place& \#pick/place & Rank pred. & Explanation\\
          top human & pred. human& human (of 8) &\\
          \hline
          12 & 12& 1 & \checkmark\\
          8 & 8 & 1 & \checkmark \\ 
        7 & 7 & 1 &  \checkmark\\ 
         6 & 6 & 1 & \checkmark \\ 
         6 & 6 & 1 & \checkmark\\ 
         6 & 6 & 1 & \checkmark \\ 
         4 & 2 & 2 & (a) \\ 
         4 & 2 & 2 & (b)\\ 
         4 & 2 & 2 & (c)\\ 
         0 & - & - & \checkmark (d) \\ 
    \end{tabular}
    \caption{Summary of results. The algorithm works especially well
      for more interactions, where it is less likely that someone else was
      also present by accident. (a) Predicted one twin, correct answer
      was the other. (b) Both twins were present in many interactions
      by coincidence, one of them was ranked first.  (d) Failure due to low signal-to-noise ratio. (d) Untouched
      object successfully gets no appearances or disappearances
      matched to a human.} 
    \label{tab:res_objects}
\end{table}
\end{footnotesize}

\section{Conclusion}
In many datasets we can find entities, subsets of the data with internal consistency, such as people in a video, popular topics from Twitter feeds, or product properties from sentences in its reviews. Currently, most practitioners wanting to find such entities use clustering.

We have demonstrated that the problem of entity finding is well-modeled as an instance of the heavy hitters problem and provided a new algorithm, \HAC, for heavy hitters in continuous non-stationary domains. In this approach, entities are specified by indicating how close data points have to be in order to be considered from the same entity and when a subset of points is big enough to be declared an entity. We proved, both theoretically and experimentally, that random sampling (on which \HAC\ is based), works surprisingly well on this problem. Nevertheless, future work on more complex or specialized algorithms could achieve better results.

We used this approach to demonstrate a home-monitoring system that allows a wide variety of post-hoc queries about the interactions among people and objects in the home.

\section{Acknowledgements}
We gratefully acknowledge support from NSF grants 1420316, 1523767 and 1723381 and from AFOSR grant FA9550-17-1-0165. F. Alet is supported by a La Caixa fellowship.  R. Chitnis is supported by an NSF GRFP fellowship. Any opinions, findings, and conclusions or recommendations expressed in this material are those of the authors and do not necessarily reflect the views of our sponsors.\\
We want to thank Marta Alet, S\'{i}lvia Asenjo, Carlota Bozal, Eduardo Delgado, Teresa Franco, Llu\'is Nel-lo, Marc Nel-lo and Laura Pedemonte for their collaboration in the experiments and Maria Bauza for her comments on initial drafts.




\newpage
\bibliographystyle{named}
\bibliography{ref.bbl}

\newpage
\appendix
\SetCommentSty{mycommfont}

\section{Appendix: proofs and detailed theoretical explanations}

\label{appendix:sec:appendix}
{
\begin{table}[h]
    \hskip-0.2cm
    \begin{tabular}{l @{\hspace{.2cm}} r @{\hspace{.2cm}} l @{\hspace{.2cm}} l}
        \multicolumn{4}{l}{\small\textbf{Thm. \ref{appendix:thm:whp} and corollary \ref{appendix:corollary:triangular} prove that with high probability:}}  \\
        {\small \textbf{All}} &{\small $(r,f)$-dense pts} & {\small will} &\multirow{2}{*}{\small have an output within $3r$}\\
        {\small\textbf{All}} & {\small $(5r, (1-\epsilon)f)$-sparse pts} & {\small won't} &\\
        \multicolumn{4}{l}{\small\textbf{For \HAC\ with radius $2r$, thm. \ref{appendix:thm:dense_likely_near} and corollary \ref{appendix:corollary:triangular} prove:}}  \\
        {\small Most} & {\small $(r,f)$-dense pts} & will &\multirow{2}{*}{\small have an output within $r$}\\
        {\small \textbf{All}} &{\small $(3r, (1-\epsilon)f)$-sparse pts} & won't &\\
        \multicolumn{4}{l}{\small\textbf{Fig. \ref{fig:guarantees} and thm \ref{appendix:thm:high-dim-inf-dist} show that in high dimensions:}}\\ 
        {\small Most} &{\tiny $((1+\Delta)r,f)$}{\small-dense pts} & {\small will} & \multirow{2}{*}{\small " " " within $(1+\Delta_2)r$}\\
        {\small Most} &{\tiny $((1+\Delta)r, (1-\epsilon)f)$}{\small-sparse pts} & {\small won't} &\\
    \end{tabular}
    \caption[\HAC\ guarantees]{\textbf{Summary of guarantees}. $\epsilon$ is a parameter of the algorithm that affects memory and runtime. Different levels of guarantees have different levels of certainty. Guarantees are constructed to be easy to verify experimentally.}
    \label{appendix:tab:guarantees}
\end{table}
}
We make a guarantee for every dense or sparse point in space, even
those that are not in the dataset. Our guarantees are probabilistic;
they hold with probability $1-\delta$ where $\delta$ is a parameter of
the algorithm that affects the memory usage. We have three types of guarantees, from loose but very
certain, to tighter but less certain. Those guarantees are summarized in table \ref{appendix:tab:guarantees}.
For simplicity, the guarantees in that table assume that there's a single radius $r_{\min}=r_{\max}=r$. We also start by proving properties of the single radius algorithm.

First we prove that if we run $HAC(f,\epsilon,\delta,2r)$ \textit{most} $(r,f)$-dense points will have an output within distance $r$ using a small amount of memory (and, in particular, not dependent of the length of the stream). Notice that, for practical values such as $f=2\%, \delta = 0.5$ we're guaranteeing that an $(r,f)$-dense point will be \textit{covered} with $99\%$ probability.

\begin{theorem}
\label{appendix:thm:dense_likely_near}
Let $\epsilon<1$. For any $(r,f)$-interesting point $p$, $HAC(f,\epsilon,\delta,2r)$ outputs a point within distance $r$ with probability $(1-f\delta)$. Moreover, it always needs at most $\Theta(\frac{\log (f\delta)d}{\epsilon f})$ memory and $\Theta(\frac{\log (f\delta)}{\epsilon f})$ time per point. Finally, it outputs at most $\Theta(\frac{\log (f\delta)}{\epsilon f})$ points.
\end{theorem}
\textbf{Proof} We maintain $m$ independent points that hop to the $t$-th point with probability $\frac{1}{t}$. They carry an associated counter: the number of points that came after its last hop and were within $2r$ of its current position. When the algorithms is asked for centers, it returns every point in memory whose counter is greater than $(1-\epsilon)fN$.

By triangular inequality any point within $p$'s ball will count towards any other point in the sphere, since we're using a radius of $2r$. Moreover, the first $\epsilon f$ points within $p$'s sphere will come before at least a fraction $(1-\epsilon)f$ of points that within $p$'s ball. Therefore there's at least a fraction $(1-\epsilon)f$ of points within distance $r$ of point $p$ that, if sampled, would be returned.

We have $m=\frac{\log{(f^{-1}\delta^{-1}})}{f\epsilon}$ samples. The probability that \textit{none} of that $(1-\epsilon)f$ fraction gets sampled is:
$$\left(1-\epsilon f \right)^m \leq e^{-\epsilon f m} = e^{-\epsilon f\frac{\log{(f^{-1}\delta^{-1}})}{f\epsilon}} = e^{\log{(f\delta})} = f\delta$$
Therefore the probability that at least one sample is within that fraction (and therefore at least there's an output within $r$ of $p$) is at least $(1-f\delta)$.
\QEDA

Now we want to prove that the same algorithm will not output points near sufficiently sparse points. 

\begin{lemma}
If we run $HAC(f, \epsilon, \delta, R)$, any $(\Delta,(1-\epsilon))$-sparse point will \textit{not} have an output point within $\Delta-R$.
\label{appendix:lemma:triangular}
\end{lemma}
Let us prove it by contradiction. Let $p$ be a $(\Delta,(1-\epsilon))$-sparse point. Suppose $HAC(f,\epsilon, \delta, R)$ outputs a point within distance $\Delta-R$ of $p$. By triangular inequality, any point within distance $R$ of the output is also within distance $\Delta$ of $p$. Since to be outputed a point has to have at least a fraction $(1-\epsilon)f$ within distance $R$ that implies there is at least a fraction $(1-\epsilon)f$ within $\Delta$ of $p$. However, this contradicts the definition that $p$ was $(\Delta, (1-\epsilon)f)$-sparse.
\QEDA

\begin{corollary}
If we run $HAC(f, \epsilon, \delta, 2r)$, any $(3r,(1-\epsilon)f)$-sparse point will not have an output within $r$ and any $(5r,(1-\epsilon)f)$-sparse point will not have an output within $3r$.
\label{appendix:corollary:triangular}
\end{corollary}
\textbf{Proof}
Use $R=2r$ and $\Delta=3r,\Delta=5r$ in the previous lemma.
\QEDA

We have shown that running $HAC(f,\epsilon,\delta,2r)$, \textit{most} $(f,r)$-dense points will have an output within $r$ and \textit{none} of the $(3r,(1-\epsilon))$-sparse will. Therefore we can use HAC as a \textit{dense}/noise detector by checking whether a point is within $r$ of an output.

We now want a probabilistic guarantee that works for \textit{all} dense points, not only for most of them. Notice there may be an uncountable number of dense points and thus we cannot prove it simply using probability theory; we need to find a correlation between results. In particular we will create a finite coverage: a set of representatives that is close to all \textit{dense} points. Then we will apply theorem \ref{appendix:thm:dense_likely_near} to those points and translate the result of those points to \textit{all} dense points.

\begin{theorem}

\label{appendix:thm:whp}
Let $\epsilon<1$. With probability $1-\delta$, for any $(r,f)$-interesting point $p$, $HAC(f,\epsilon,\delta,2r)$ outputs a point within distance $3r$. Moreover, it always needs at most $\Theta(\frac{\log (f\delta)d}{\epsilon f})$ memory and $\Theta(\frac{\log (f\delta)}{\epsilon f})$ time per point. Finally, it outputs at most $\Theta(\frac{\log (f\delta)}{\epsilon f})$ points.
\end{theorem}
\textbf{Proof} Let $D$ be the set of $(r,f)$-dense points. Let $D^*=\{p_1,p_2,\dots\}$ be the biggest subset of $D$ such that $B(p_i,r)\cap B(p_j,r) = \emptyset$ for any $i\neq j$. Since the pairwise intersection is empty and $|B(p_i,r)|\geq fN$ for any $i$, we have $|\bigcup_i B(p_i,r)| = \sum_i |B(p_i,r)| \geq |D^*|\cdot fN$. However, $N\geq |\bigcup_i B(p_i,r)|$, so we must have $|B^*| \leq \frac{1}{f}$. 

We now look at a single run of $HAC(f,\epsilon, \delta, 2r)$. Using theorem \ref{appendix:thm:dense_likely_near}, for any $p_i\in D^*$ the probability of having a center within $r$ is at least $1-\delta f$. Therefore, by union bound the probability that \textit{all} $p_i\in D^*$ have a center within $r$ is at least: $1-\delta f\frac{1}{f} = 1-\delta$.

Let us assume that all points in $D^*$ have an output within $r$. Let us show that this implies something about \textit{all} dense points, not just those in the finite coverage. For any point $p\notin D^* \exists p_i\in D^*$ s.t. $B(p,r)\cap B(p_i,r) \neq \emptyset$. If that were not the case, we could add $p$ to $D^*$, contradicting its maximality. Since their balls of radius $r$ intersect this implies their distance is at most $2r$. We now know $\exists p_i\in D^*$ s.t. $d(p,p_i)\leq 2r$ and that $\exists$ center $c$ s.t. $d(c,p_i)\leq r$. Again by triangular inequality, point $p$ will have a center within distance $3r$.

Both runtime and memory are directly proportional to the number of samples, which we specified to be $m=\frac{\log{(f^{-1}\delta^{-1}})}{f\epsilon}$.
\QEDA

Let us now move to the multiple radii case. For that we need the following definition:
\begin{definition}
$r_f(p)$ is the smallest $r$ s.t. $p$ is $(r,f)$-dense. For each point $p$ we refer to its \textit{circle/ball} as the sphere of radius $r_f(p)$ centered at $p$.
\end{definition}
Note that now \textit{any} point will be \textit{dense} for some $r$.
Given that all points are dense for some $r$, there are two ways of giving guarantees:
\begin{itemize}
    \item All output points are paired with the radius needed for them to be dense. Then, guarantees can be made about outputs of a specific radius.
    \item We can still have a $r_{\max}$, for which all guarantees for the single radius case apply directly.
\end{itemize}

When we pair outputs with radius we call \textit{an output of radius $r$} to an output that needed a radius $r$ to be dense. In that case, we can make a very general guarantee about not putting centers near sufficiently sparse regions, where sparsity is a term relative to $r_f$.
\begin{lemma}
If we run $HAC(f, \epsilon, \delta)$; for \textit{any} point $p$, there will not be an output $o$ of radius $r(o)$ within distance less than $r_{(1-\epsilon)f}(p)-r(o)$.
\label{appendix:lemma:general_triangular}
\end{lemma}
\textbf{Proof} Similar to \ref{appendix:lemma:triangular}, we can assume there is an output point within that distance and apply triangular inequality. We then see that all points within distance $r(o)$ of the output would be within distance $r_{(1-\epsilon)f}(p)$ of $p$. However, we know that the output has at least a fraction $(1-\epsilon)f$ within distance $r(o)$, contradicting the minimality of $r_{(1-\epsilon)f}(p)$.
\QEDA

\begin{theorem}
For any tuple $(\epsilon<1, \delta, f, \gamma > 1)$, for any point $p$ s.t. $r_f\leq \frac{r_{\max}}{2\gamma}$ our algorithm will give an output point within $r_f(p)$ of at most radius $2\gamma r_f(p)$  with probability at least $1-\delta f$.\\
Moreover, the algorithm always needs at most $\Theta\left(\frac{\log (f\delta)}{\epsilon f}\log_\gamma\left(\frac{r_{\max}}{r_{\min}}\right)\right)$ memory and $\Theta(\frac{\log (f\delta)}{\epsilon f})$ time per point. Finally, it outputs at most $\Theta\left(\frac{\log (f\delta)}{\epsilon f}\right)$ points.
\label{appendix:theorem:dense_likely_near_multiple_radii}
\end{theorem}
\textbf{Proof} Let us run our algorithm with multiple radius and then filter only the outputs of radius less than $2\gamma r_f(p)$. Since radius are discretized we are actually filtering by the biggest radius of the form $r_0\gamma^c$. Nevertheless, since there's one of those radii for every $\gamma$ scale, there must be one between $2r_f(p)$ and $2\gamma r_f(p)$, let's call it $r'$. Running the multiple radii version then filtering by $r'$ is equivalent to running the single radius version with radius $r'$. Since $r'\geq 2r_f(p)$, counters for $r'$ must all be at least as big as for $2r_f(p)$ and thus the outputs for $r'$ are a superset of those for $2r_f(p)$. We can apply the equivalent theorem for a single radius (thm \ref{appendix:thm:dense_likely_near}) to know that if we had run the single radius version $HAC(f,\epsilon,\delta,2r_f(p))$ we would get an output within $r_f(p)$ with probability at least $1-\delta f$. Therefore the filtered version of multiple radii must also do so. Since we have filtered at least an output of radius less than $r'\leq 2\gamma r_f(p)$ within distance $r_f(p)$ that means the multiple radii version will output such a center with probability at least $1-\delta f$.

Since memory mainly consists of an array of dimensions $m=\frac{\log(f^{-1}\delta^{-1})}{f\epsilon}, c=\log_\gamma\frac{r_{\max}}{r_{\min}}$, the memory cost is $\Theta\left(\frac{\log (f\delta)}{\epsilon f}\log_\gamma\left(\frac{r_{\max}}{r_{\min}}\right)\right)$. Notice that, to process a point we do not go over all discrete radii but rather only add a counter to the smallest radius that contains it, therefore the processing time per point is $\Theta(m) = \Theta\left(\frac{\log (f\delta)}{\epsilon f}\right)$.
\QEDA

\begin{theorem}
For any tuple $(\epsilon<1, \delta, f, \gamma > 1)$, with probability $1-\delta$, for any point $p$ s.t. $r_f\leq \frac{r_{\max}}{2\gamma}$ our algorithm will give an output point within $3r_f(p)$ of at most radius $2\gamma r_f(p)$.\\
Moreover, the algorithm always needs at most $\Theta\left(\frac{\log (f\delta)}{\epsilon f}\log_\gamma\left(\frac{r_{\max}}{r_{\min}}\right)\right)$ memory and $\Theta(\frac{\log (f\delta)}{\epsilon f})$ time per point. Finally, it outputs at most $\Theta\left(\frac{\log (f\delta)}{\epsilon f}\right)$ points.
\label{appendix:theorem:whp_multiple_radii}
\end{theorem}
\textbf{Proof} The exact same reasoning of a finite coverage of theorem \ref{appendix:thm:whp} can be applied to deduce this theorem from theorem \ref{appendix:theorem:dense_likely_near_multiple_radii} changing $r$ to $r_f(p)$.
\QEDA

Notice how we can combine lemma \ref{appendix:lemma:general_triangular}, that proves that sparse enough points will not get an output nearby, with theorems \ref{appendix:theorem:dense_likely_near_multiple_radii}, \ref{appendix:theorem:whp_multiple_radii} to get online dense region detectors with guarantees.

Note that we proved guarantees for \textit{all} points and for \textit{all} possible metrics. Using only triangular inequality we were able to get reasonably good guarantees for a non-countable amount of points, even those not on the dataset. We finally argue that the performance of \HAC\ in high dimensions is guaranteed to be almost optimal.
\subsection[Stronger guarantees in high dimensions]{The \textit{blessing} of dimensionality: stronger guarantees in high dimensions}
\label{appendix:sec:high-dim}
\paragraph{Intuition}
In high dimensions many clustering algorithms fail; in contrast, our performance can be shown to be provably good in high dimensions. We will prove asymptotically good performance for dimension $d\rightarrow \infty$ with a convergence fast enough to be meaningful in real applications. In particular, we will prove the following theorem:
\newtheorem*{theorem*}{Theorem}
\begin{theorem*}
Let $\kappa = \frac{4}{c^2}e^{-c^2/4}, l=e^{-\beta}/(1-\kappa),\hat{f} = f/(1-2\kappa)$, $\Delta_1 = c\sqrt{\frac{2}{d}} + O(d^{-1})$, $\Delta_2 = 2c\sqrt{\frac{2}{d}} + O(d^{-1})$.
Let $m$ $d$-dimensional samples come from Gaussians $Z_1,\dots, Z_k$ with means infinitely far apart, unit variance and empirical frequencies $f_1,\dots, f_k$. If we run \HAC\ with radius $r=U+2c=\sqrt{2}\sqrt{d+2\sqrt{d\beta}+2\beta}+2c$ and frequency $f$, any point $p$ with $r_{\hat{f}}(p)\leq r$ will have an output $o$ within $(1+\Delta_1)r_f(p)$ with associated radius ($r$) at most $(1+\Delta_2)r_f(p)$ with probability at least $\left(1-\kappa-\delta f - e^{-fn/3}\left(e^\kappa+e^l\right)\right)$.

Moreover, the probability that a point $p$ has $r_{(1-\epsilon)(1-2l)f}(p) > r$ yet has an output nearby is at most $\left(\kappa+e^{-fn/3}\right)$
\end{theorem*}
Later, we will add 2 conjectures that make guarantees applicable to our experiments. Since the proof is rather long, we first give a roadmap and intuition.

If we fix a point $p$ in Gaussian $Z_k$ we can look at other points $q\sim Z_k$ and their distance to $p$, we call this distribution $dist_p$. $r_f(p)$ is the distance for which a fraction $f$ of the dataset is within $r_f(p)$ of $p$. Since all but $f_k$ of points are infinitely far away; $r_f(p)$ is equivalent to the $f/f_k$ quantile of $dist_p$. One problem is that this quantile is a random variable; which we will have to bound probabilistically.
\begin{figure}
    \centering
    \includegraphics[width=\linewidth]{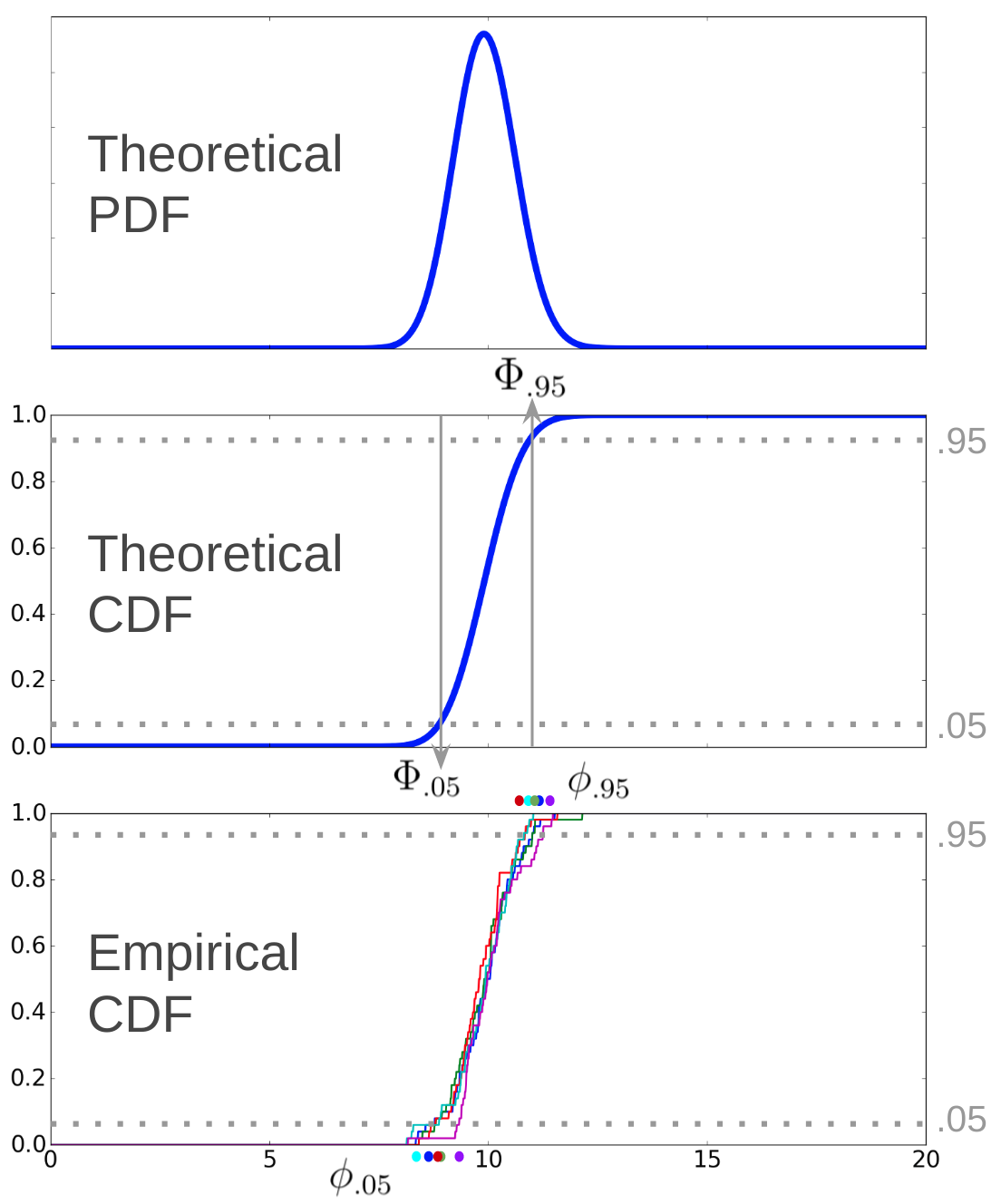}
    \caption[Concentrated PDF leads to bounds on empirical quantiles]{Fixing a point $p$, $dist_p(q)=|q-p|, q\sim Z_k$ is a random variable that, for high dimensions, is well concentrated. If $f/f_k=x$ then $r_f(p)=\phi_x(p)$, the empirical quantile (colored dots in the bottom figure). \\
    We want to prove that empirical quantiles $\phi_x$ are pretty close to one another, which would imply that all points have very similar $r_f(p)$. For example, in this case all samples from $\phi_x$ for $0.05\leq x \leq 0.95$ are inside [8,12]. \\Our proof will first look at the theoretical quantiles $\Phi_x(p)$ and then bound $\phi_x(p)$.
    }
    \label{appendix:fig:pdf_cdf_experimental}
\end{figure}
Remember that quantile $x$ of the \textit{theoretical} distribution is simply the inverse of the Cumulative Density Function; i.e. there is a probability $x$ that a sample is smaller than the $x$ quantile. We denote the quantile for $dist_p$ by $\Phi_x(p)$, sometimes omitting $p$ when implicit; notice $\Phi_x(p)$ is a function. For finite data, samples don't follow the exact CDF and therefore quantiles are random variables; we denote these empirical quantiles by $\phi_x(p)$. We refer to figure \ref{appendix:fig:pdf_cdf_experimental} for more intuition.

\begin{enumerate}
    \item Model the data as a set of $d$-dimensional Gaussians with the same variance $\sigma^2\cdot Id$ but different means. If we want to have uniform noise, we can have many Gaussians with only 1 sample.
    \item Without loss of generality (everything is the same up to scaling) assume $\sigma=1$.
    \item Most points in a high dimensional Gaussian lie in a shell between $\sqrt{d-1}-c$ and $\sqrt{d+1}+c$, for a small constant $c$ (lemma \ref{appendix:lemma:online_pdf}). We will restrict our proof to points $p$ in that shell.
    \item The function we care about, $dist_p$, from a particular fixed point $p$ to points coming from the same Gaussian follows a \textit{non-central chi distribution}, a complex distribution with few known bounds, we will thus try to avoid using it.
    \item The distribution $dist(p,q)^2$ where $p,q\sim N(0,1)$ follows a (central) chi-squared distribution, $\chi_d^2$, a well studied distribution with known bounds.
    \item $dist(p,q)^2$ where $p,q\sim N(0,1)$ and $dist(p,q)^2$ where $p,q\sim N(0,1), p\in \text{shell}$ are very similar distributions because most $p\sim N(0,1)$ are in the shell. Bounds on the former distribution will imply bounds on the latter.
    \item We need to fix $p$ and only sample $q$. We show quantiles of the distribution are 1-Lipschitz and use it along with Bolzano's Theorem to get bounds with fixed $p\in$ shell.
    \item Since we care about finite-data bounds we need to get bounds on empirical quantiles, we bridge the gap from theoretical quantiles using Chernoff bounds.
    \item We will see that quantiles are all very close together because in high dimensional Gaussians most points are roughly at the same distance. For any point $p$ we will be able to bound its radius $r_f(p)$ using the bounds on quantiles of the distance function. 
    \item With this bound we will be able to bound the ratio between the radius of a point $p$ and the distance to its closest output or the radius of such output.
    \item We join all the probabilistic assertions made in the previous steps via the union bound, getting a lowerbound for all the assertions to be simultenously true.
\end{enumerate}
\paragraph{Proof}
In high dimensions, Gaussians look like high dimensional shells with all points being roughly at the same distance from the center of the cluster, which is almost empty. We will first assume Gaussians are infinitely far away and Gaussians of variance 1. For many lemmas we will assume mean 0 since it doesn't lose generality for those proofs.

We first use a lemma 2.8 found in an online version of \cite{blum2016foundations} \footnote{\url{https://www.cs.cmu.edu/~venkatg/teaching/CStheory-infoage/chap1-high-dim-space.pdf}}, which was substituted by a weaker lemma in the final book version. This lemma formalizes the intuition that most points in a high dimensional Gaussian are in a shell:
\begin{lemma}
\label{appendix:lemma:online_pdf}
For a $d$-dimensional spherical Gaussian of variance 1, a sample $p$ will be outside the shell $\sqrt{d-1}-c\leq |p| \leq \sqrt{d-1}+c$ with probability at most $\frac{4}{c^2}e^{-c^2/4}$ for any $c>0$.
\end{lemma}
We will prove that things work well for points inside the shell; which for $c=3$ it's $95\%$ of points and $c=4$ it's $99.6\%$. For future proofs let us denote $\kappa(c)=\frac{4}{c^2}e^{-c^2/4}$; to further simplify notation we will sometimes omit the dependence on $c$.
\begin{lemma}
\label{appendix:lemma:kappa}
Let $dist(p,q)=|q-p|$ with $p,q\sim N(0,1)$ and $|p|\in[\sqrt{d-1}-c,\sqrt{d-1}+c]$, but no restriction on the norm of $q$. Then: 
$$Prob\left(dist(p,q)\leq \sqrt{2}\sqrt{d+2\sqrt{\beta\cdot d}+2\beta}\right)\leq \frac{e^{-\beta}}{\kappa}$$ and $$Prob\left(dist(p,q)\leq \sqrt{2}\sqrt{d+2\sqrt{\beta\cdot d}+2\beta}\right)\leq \frac{e^{-\beta}}{1-\kappa}$$
\end{lemma}
\textbf{Proof} If we forget for a moment about the shell and consider $a,b\sim N(0,1)$ then $(a-b) \sim N(0,\sqrt{2})$ and $dist(a,b)^2=|a-b|^2 \sim 2\chi_d^2$.

\begin{figure}
    \centering
    \includegraphics[width=.5\textwidth]{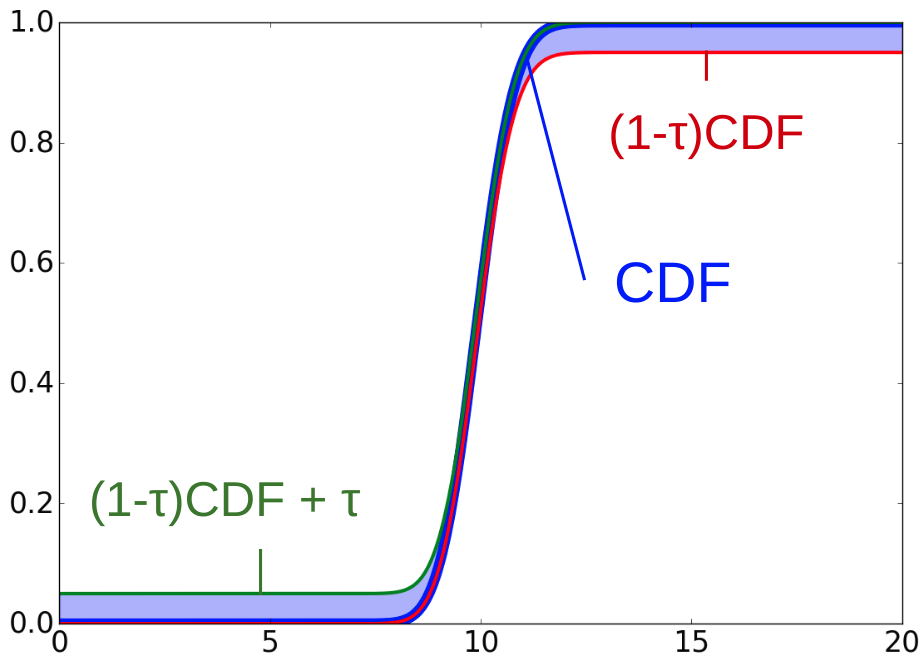}
    \caption[Close Cumulative Density Functions]{The CDF for $p\in shell$ (in deep blue) creates a small interval (light blue) for the CDF for unbounded $p$.}
    \label{appendix:fig:scaled_cdf_tau}
\end{figure}

We now observe that there are two options for $a$, either it is inside the shell ($|a|\in [\sqrt{d-1}-c, \sqrt{d-1}+c]$) or outside. Since the probability of being inside the shell is very high, $dist(a\in shell,b)$ and $dist(a,b)$ are very close. In the worst case, using that $a$ and $b$ are chosen independently, we have:
\begin{align*}
 CDF[dist(a, b)] = &(1-\kappa) \cdot CDF[dist(a\in shell,b)]\\ 
 &+ \kappa \cdot CDF[dist(a\notin shell, b)]  
\end{align*}
using $0\leq CDF[dist(a\notin shell, b)] \leq 1$ we can get the following inequalities:
\begin{align*}
& CDF[dist(a\in shell,b)] (1-\kappa) \leq  CDF[dist(a,b)] \\
\kappa + & CDF[dist(a\in shell,b)] (1-\kappa) \geq CDF[dist(a,b)]    
\end{align*}
$$P\left(dist(a\in shell, b)\leq y\right) \leq \frac{P\left(dist(a,b)\leq y\right)}{1-\kappa}$$
$$P\left(dist(a\in shell, b)\geq y\right) \leq \frac{P\left(dist(a,b)\geq y\right)}{1-\kappa}$$

Now \cite{laurent2000adaptive} shows that:

\begin{align*}
&P\left(\chi_d^2\leq  d-2\sqrt{\beta d} \right) &\leq e^{-\beta}\\
&P\left(\chi_d^2\geq  d+2\sqrt{\beta d}+2\beta\right) &\leq e^{-\beta}    
\end{align*}

Remember that $dist(a,b)\sim\sqrt{2}\chi_d$, we transform $\chi^2_d$ into $\sqrt{2}\chi_d$ by taking the square root and multiplying by $\sqrt{2}$:

\begin{align*}
&P\left(dist(a,b)\leq \sqrt{2} \sqrt{d-2\sqrt{d\beta}}\right) &\leq e^{-\beta} \hfill \\
&P\left(dist(a,b)\geq \sqrt{2} \sqrt{d+2\sqrt{d\beta}+2\beta}\right) &\leq e^{-\beta} \hfill
\end{align*}
To shorten formulas let us denote the lowerbound by $L=\sqrt{2} \sqrt{d-2\sqrt{d\beta}})$ and the upperbound by $U=\sqrt{2} \sqrt{d+2\sqrt{d\beta}+2\beta}$.
Finally, if we look at the $e^{-\beta}$ and $(1-e^{-\beta})$ quantiles we know from the equations above that they must be above $L$ and below $U$.

Note that setting $\beta=3$ we get bounds on quantiles $5\%,95\%$ and setting $\beta=4$ we get bounds on quantiles $2\%,98\%$.

We now have bounds on \textit{theoretical} quantiles for $dist(a,b)$; as mentioned before we can translate them to bounds on $dist(a\in shell, b)$ getting probabilities bounded by $l=\frac{e^{-\beta}}{1-\kappa}$.

\QEDA

Up until now we have proved things about arbitrary $a,b\sim N(0,1)$. Our ultimate goal is proving that the radius for a particular point $p$ in the shell cannot be too big or too small. To reflect this change in goal we change the notation from $a,b$ to $p,q$. $r_f(p)$ is defined as the minimum distance for a fraction $f$ of the dataset to be within distance $r$ of $p$. Therefore we care about samples from $dist(p,q)$ \textit{with constant $p$}. Since $p$ is sampled only once those samples are correlated and we have to get different bounds.

\begin{lemma}
Let $p, |p|\in[\sqrt{d-1}-c,\sqrt{d-1}+c]$ be fixed. Let $dist_p$ be the \textit{theoretical} $dist(p,q), q\sim N(0,1)$. Then the quantiles $l$ and $1-l$ are both contained in $[L-2c, U+2c]$.
\label{appendix:lemma:theoretical_quantiles}
\end{lemma}

\textbf{Proof} From the previous lemma \ref{appendix:lemma:kappa} we know that when $p$ is \textit{not} fixed, the quantiles $l$ and $1-l$ from that distribution are in $[L, U]$.

By rotational symmetry of the Gaussian we know that this distribution only depends on the radius $|p|$; overriding notation let us call it $dist_{|p|}$.

Let us now consider two radius $r$, $r'$. We can consider the path from $p, |p|=r$ to $q\sim N(0,1)$ passing through $p' = p\cdot \frac{r'}{r}$, which upperbounds the distance from $p$ to $q$ by triangular inequality. The shortest path from $p$ to $p'=p\cdot \frac{r'}{r}$ is following the line from $p$ to the origin taking length $|r-r'|$.

We thus have that $|dist(p,q)-dist(p\cdot \frac{r'}{r})|\leq |r-r'|$ and thus the Cumulative Density Function of $dist_{r}$ is upperbounded by $dist_{r'}$ shifted by $|r-r'|=dist(p,p')$.

\begin{figure}
    \centering
    \includegraphics[width=\linewidth]{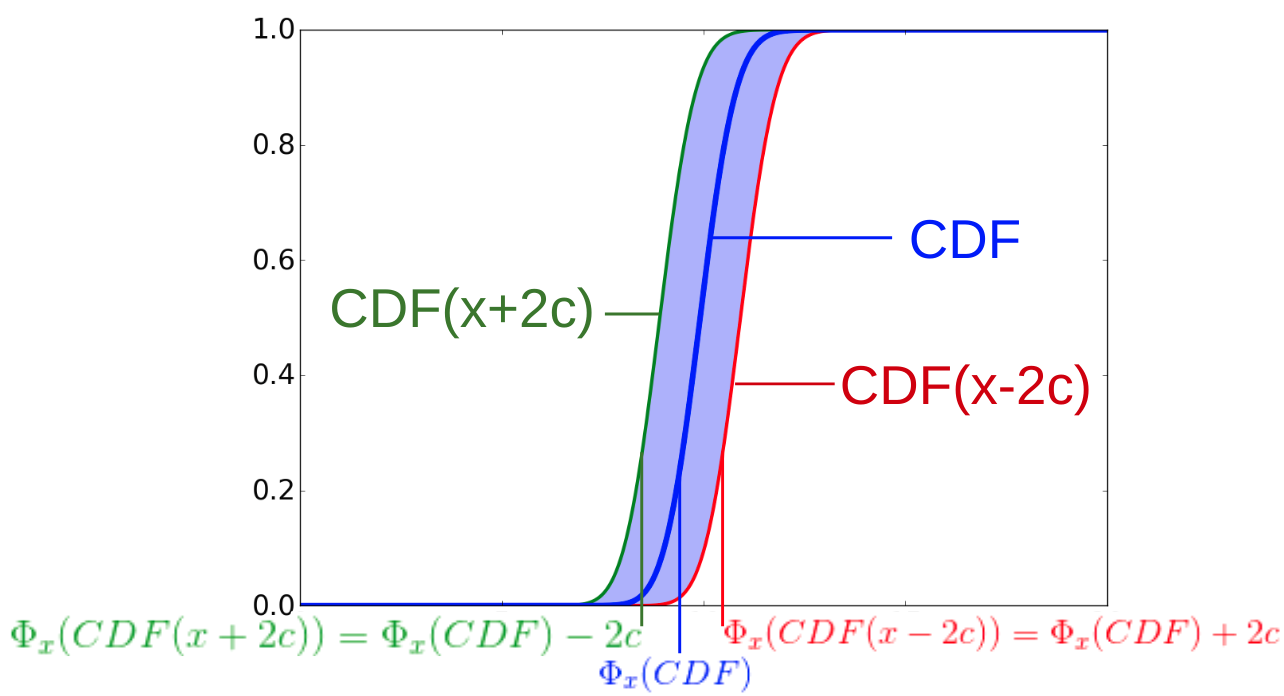}
    \caption[Shifting the CDF translates its quantiles]{Shifting the CDF simply adds a factor $2c$ or $-2c$ to its quantiles}
    \label{appendix:fig:shifted_cdf}
\end{figure}

As figure \ref{appendix:fig:shifted_cdf} illustrates, this implies that the quantiles of $dist_{|p|}$ are $1$-Lipschitz and, in particular, also continuous. Remember that a function $f(x)$ is $1$-Lipschitz if $|f(x)-f(y)| \leq |x-y|$.

Since $p$ and $q$ are chosen independently, we can first select $p$ then $q$. Let us consider three options:
\begin{enumerate}
    \item $\Phi_x(r) < \Phi_x(p\in \text{shell}) \forall r\in[\sqrt{d-1}-c, \sqrt{d-1}+c]$. Taking the lower $x$ fraction for every $p$ represents fraction $x$ of the total samples $(p,q)$. We have data of fraction $x$ all less than $\Phi_x(p\in shell)$. This contradicts the definition of quantile. 
    \item $\Phi_x(r) > \Phi_x(p\in \text{shell}) \forall r\in[\sqrt{d-1}-c, \sqrt{d-1}+c]$. By definition of $\Phi_x(r)$ no other sample can be below $\Phi_x(p\in \text{shell})$ which implies that the $x$ quantile is above $\Phi_x(p\in shell)$. Again this contradicts the definition of quantile.
    \item $\exists r_1 \text{ s.t. } \Phi_x(r_1) \leq \Phi_x(p\in \text{shell}) \text{ and } \exists r_2 \text{ s.t. } \Phi_x(r_2) \geq \Phi_x(p\in \text{shell})$. Since $\Phi_x(r)$ is continuous, by Bolzano's Theorem we know: $$\exists r_{Bolzano}(x) \text{ s.t. } \Phi_x(r_{Bolzano}(x))=\Phi_x(p\in \text{shell})$$
\end{enumerate}
\begin{figure}
    \centering
    \includegraphics[width=\linewidth]{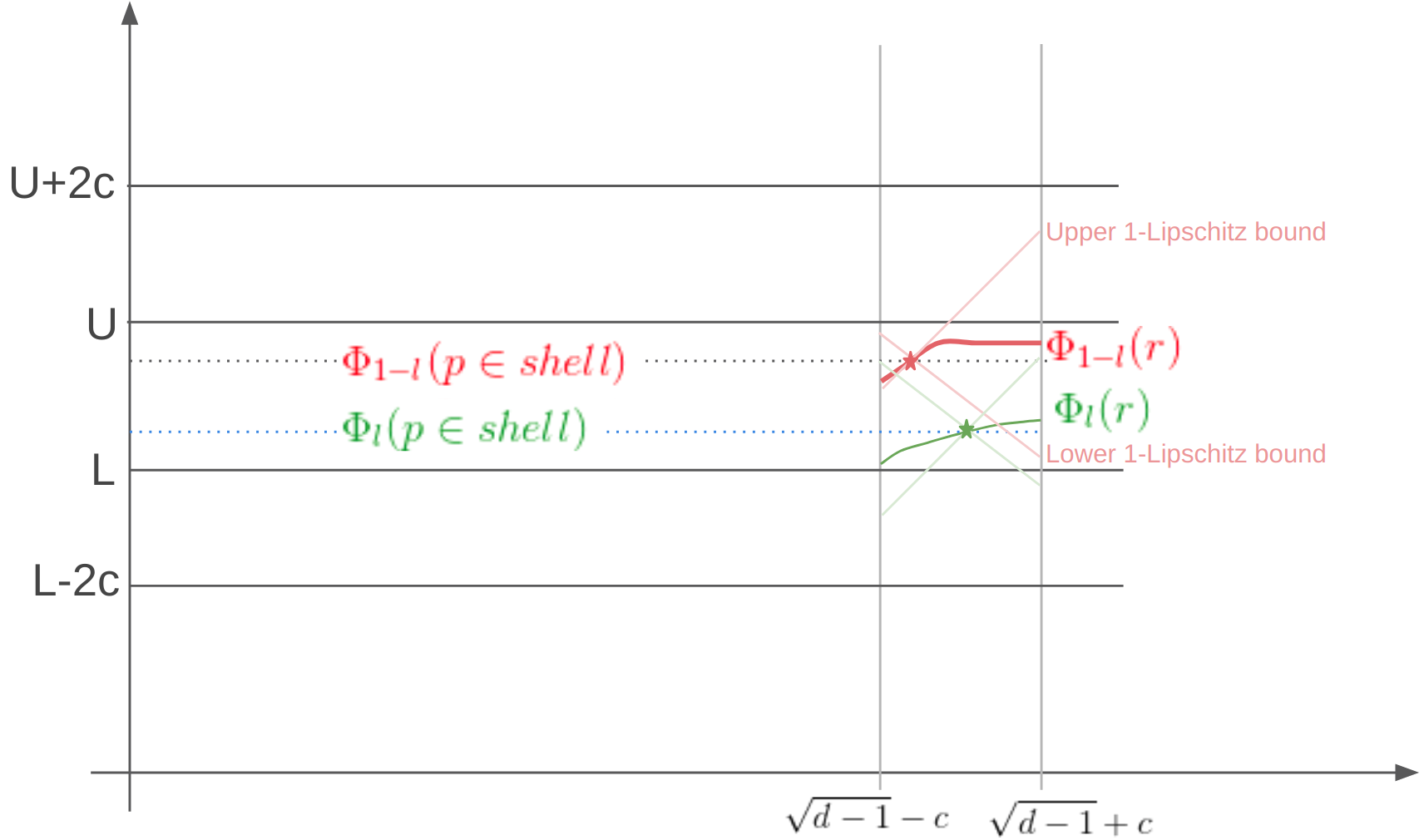}
    \caption[Using 1-Lipschitzness and Bolazano's theorem]{Bolzano's Theorem guarantees there's a point ({\color{red}$\star$},{\color{OliveGreen}$\star$} in the figure) where $\Phi_x(p\in shell)=\Phi_x(r_{Bolzano}(x))$. From there we use that $\Phi_x(r)$ is 1-Lipschitz to delimit a cone for all points in the shell.}
    \label{appendix:fig:lipschitz}
\end{figure}
After this, as shown in figure \ref{appendix:fig:lipschitz}, we apply that quantiles are 1-Lipschitz and since the maximum distance in that interval is $(\sqrt{d-1}+c)-(\sqrt{d-1}-c)=2c$ we know that for all points in the shell their theoretical quantiles $l$ and $1-l$ must be inside $[L-2c, U+2c]$.
\QEDA

Note that we now have bounds on \textit{theoretical} quantiles; \textit{empirical} quantiles (those that we get when the data comes through) will be noisier for finite data and thus quantiles are a bit more spread; as illustrated in figure \ref{appendix:fig:pdf_cdf_experimental}. This difference can be bounded with Chernoff bounds. In particular let us compare the probability that the empirical $2l$ and $(1-2l)$ quantiles are more extreme than the theoretical $l$ and $(1-l)$ quantiles.

\begin{lemma}
Let us have $p$ fixed s.t. $|p|\in[\sqrt{d-1}-c,\sqrt{d-1}+c]$ and take $m$ samples $q_{1:m}\sim N(0,1)$. Then with probability higher than $1-2e^{-lm/3}$ the empirical quantile $2l$ of $\left[dist(p,q_1), dist(p,q_2),\dots, dist(p,q_m)\right]$ is bigger than $L-2c$ and the empirical quantile $1-2l$ is smaller than $U+2c$.
\label{appendix:lemma:finite-samples-chernoff}
\end{lemma}
\textbf{Proof}
Since we now have fixed $p$, we will drop it to simplify the notation.

From lemma \ref{appendix:lemma:theoretical_quantiles} we know $\Phi_{l}\geq L-2c$, $\Phi_{1-l}\leq U+2c$. We want to prove $\phi_{2l}\geq \Phi_{l} \geq L-2c$ and $\phi_{1-2l}\leq \Phi_{1-l} \leq U+2c$ with probability bigger than $1-2e^{lm/3}$.

We can bound the difference between \textit{empirical} and \textit{theoretical} quantiles of the same distribution using Chernoff bounds. The bounds on the high and low quantiles are proven in the exact same way. Let us prove it only for the lower one. 

Let $X_i = [[dist(p\in shell, q_i)\leq \phi_l]]$ where $[[]]$ is the Iverson notation; being 1 if the statement inside is true and 0 if false. Chernoff tells us that if we have independent random variables taking values in $\{0,1\}$ (as in our case) Then: $$P\left(\sum X_i\geq(1+\omega)\mu\right)\leq e^{-\omega^2\mu/3}, \mu = \mathbb{E}\left[\sum X_i\right]$$

Applying it to our case:
$$\mu = \mathbb{E}\left[\sum X_i \right] = \mathbb{E}\left[\sum_{q_i} [[dist(p\in shell, q_i)\leq \phi_l]]\right]= l\cdot m$$
$$P\left(\sum [[dist(p\in shell, q) \geq (1+\omega) \mu \right) \leq e^{-\omega^2\mu/3}$$
Using $\mathbb{E}\left[\sum_q [[dist(p\in shell, q)\leq \phi_l]]\right]= lm$ and setting $\omega=1$ we get:
$$P\left(\sum_q [[dist(p\in shell, q)\leq \phi_l]] \geq 2lm \right) \leq e^{-lm/3}$$

By definition of empirical quantile $\phi_{2l}$ and theoretical quantile $\Phi_l$, $\phi_{2l}\leq \Phi_l \iff \sum_q [[dist(p\in shell, q)\leq \phi_l]] \geq 2lm$, which is the event whose probability we just bound. Therefore, we proved that the probability of being smaller than $\phi_l$ is bounded by $e^{-lm/3}$. Since $\phi_l\geq L-2c$ we know that the probability of the quantile being lower than $L-2c$ is even smaller than $e^{-lm/3}$.

The exact same reasoning proves that $P\left(\Phi_{1-2l}\geq U+2c\right)\leq e^{-lm/3}$. By union bound we know the probability of $\Phi_{2l}\geq L-2c$ and $\Phi_{1-2l}\leq U+2c$ happening at the same time is at least $1-2e^{-lm/3}$.
\QEDA

We are now ready to attack the main theorem. We will model the data coming from a set of Gaussians with centers \textit{infinitely} far away and \textit{empirical} frequencies $f_i$. Note that this model can model pure noise, by having many Gaussians with only 1 element sampled from them.

\begin{theorem}
\label{appendix:thm:high-dim-inf-dist}
Let $\kappa = \frac{4}{c^2}e^{-c^2/4}, l=e^{-\beta}/(1-\kappa),\hat{f} = f/(1-2\kappa)$, $\Delta_1 = c\sqrt{\frac{2}{d}} + O(d^{-1})$, $\Delta_2 = 2c\sqrt{\frac{2}{d}} + O(d^{-1})$.
Let $m$ $d$-dimensional samples come from Gaussians $Z_1,\dots, Z_k$ with means infinitely far apart, unit variance and empirical frequencies $f_1,\dots, f_k$. If we run \HAC\ with radius $r=U+2c=\sqrt{2}\sqrt{d+2\sqrt{d\beta}+2\beta}+2c$ and frequency $f$, any point $p$ with $r_{\hat{f}}(p)\leq r$ will have an output $o$ within $(1+\Delta_1)r_f(p)$ with associated radius ($r$) at most $(1+\Delta_2)r_f(p)$ with probability at least $\left(1-\kappa-\delta f - e^{-fn/3}\left(e^\kappa+e^l\right)\right)$.

Moreover, the probability that a point $p$ has $r_{(1-\epsilon)(1-2l)f}(p) > r$ yet has an output nearby is at most $\left(\kappa+e^{-fn/3}\right)$
\end{theorem}
\textbf{Proof}

\textbf{First part: $r_{\hat{f}}(p) \leq r$}\\

We will make a set of probabilitstic assertions and we will finally bound the total probability using the union bound. 

\underline{First assertion}: point $p$ belongs to the shell of its Gaussian, which we donte $Z_k$.

In a lemma \ref{appendix:lemma:kappa} we defined $\kappa$ as an upperbound on the probability of a point being inside the shell of a Gaussian. However our point $p$ is not just a random point since we know its radius is bounded by $U+2c$. We have to bound the \textit{posterior} probability given that information. In the worst case, all points outside the shell do satisfy $r_{\hat{f}}(p)\leq U+2c$. In lemma \ref{appendix:lemma:finite-samples-chernoff} we lowerbounded the probability of a point inside the shell to satisfy $r_{\hat{f}}(p)\leq U+2c$ by $1-e^{-lm_k/3}$ where $m_k$ is the number of elements in the Gaussian $Z_k$, in this case $m_k=f_kn$. Thus the posterior probability is:
\begin{align*}
\frac{(1-e^{-lf_kn/3})(1-\kappa)}{(1-e^{-lf_kn/3})(1-\kappa)+\kappa} \geq &\frac{(1-e^{-lf_kn/3})(1-\kappa)}{1\cdot(1-\kappa)+\kappa} \\
= & \left(1-e^{-lf_kn/3}\right)(1-\kappa)    
\end{align*}

\underline{Second assertion}: the shell of $Z_k$ constains at least $fn$ points.

We know $r_{\hat{f}}(p) \leq r$. Since Gaussians are infinitely far away, all points near $p$ must come from $Z_k$. This implies: $$f_k\geq \hat{f}= f/(1-2\kappa)$$

By definition of $\kappa$ we know that the probability of a sample from a Gaussian being outside its shell is $\kappa$. Applying Chernoff bounds on the amount of points outside the shell we get: 

\begin{align*}
  P\left(\frac{\#\text{points outside shell}}{n\cdot f_k}>2\kappa\right) \leq  &e^{-1^2\mu/3} = e^{-\kappa f_k} \\
  \leq  e^{-\kappa fn/(3(1-2\kappa))} \leq & e^{-\kappa fn/3}  
\end{align*}

Since we expect a fraction at most $\kappa$ that means that the amount of points inside the shell is at least $(1-2\kappa)f_k\geq (1-2\kappa)\hat{f}=(1-2\kappa)\frac{f}{1-2\kappa}=f$ with probability at least $(1-e^{-\kappa fn/3})$.

\underline{Third assertion}: given the second assertion, a point $o$ in the Gaussian will be an output.

We know we have at least a fraction $f$ of the total dataset in the shell of $Z_k$. From lemma \ref{appendix:lemma:finite-samples-chernoff}, we also know that each point $q$ in the Gaussian has a chance at least $1-e^{-e^\beta l\hat{f}n/3}$ of having $r_f(q)\leq U+2c$. Note that this guarantee was for a Gaussian from which we knew nothing. However, we know that a point already satisfies this condition; which makes it even more likely; which allows us to still use this bound.

We know the probability of each point having a small $r_f(q)$. However, we don't know how $q$ satisfying $r_f(q)\leq r$ affects $q'$ satisfying $r_f(q')\leq r$.

We compute the worst case for \HAC\ to get a lowerbound on the probability of success. In particular note that we will have a distribution over $2^{|\text{elts in the shell}|}$ states, with the $i$-th bit in each state corresponding to whether the $i$-th element in the shell had a small radius. Note that this distribution is conditioned to satisfy that the probability of the $i$-th bit to be true has to be at least $1-e^{-e^\beta \hat{f}n/(3(1-\kappa))}$. Moreover we know that the probability of \HAC\ failing (not picking any good element) is:
$$\left(1-\frac{|\text{elements }q\text{ in the shell s.t. }r_f(q)\leq r|}{n}\right)^m$$
where $m$ is the memory size.

It is easy to see that the best way to maximize this quantity under constraints is to only have the most extreme cases: either all $q$ don't satisfy this property or all do. This is because as more points $q$ satisfy the property the algorithm chances of success increase with diminishing returns.

Knowing the worst case, we can now get a lowerbound: with probability $e^{-l \hat{f}n/3}$ no $q$ is good and \HAC 's chances of success are 0. With probability $1-e^{-l \hat{f}n/3}$ we are in the usual case of a fraction $f$ of the dataset and \HAC 's chances of success are lowerbounded by:
$$1 - \left(1-f\right)^m\geq \delta f$$
where $\delta$ is the delta coming from \HAC 's guarantees.
Therefore the lowerbound for \HAC 's success is:
\begin{align*}
\left(1-e^{-l \hat{f}n/3}\right)(1-\delta f)= &\left(1-e^{-e^\beta \hat{f}n/(3(1-\kappa))}\right) (1-\delta f) \\
\geq &1 - e^{-e^\beta \hat{f}n/(3(1-\kappa))} - \delta f
\end{align*}
\underline{Bounding the total probability}

The probability of failure of the first assertion is bounded by:
\begin{align*}
1-\left(1-e^{-lf_kn/3}\right)(1-\kappa) &=\kappa+e^{-lf_kn/3}-e^{-lf_kn/3}\kappa \\
&\geq \kappa+e^{-lf_kn/3}    
\end{align*}
The probability of failure of the second assertion is bounded by $e^{-\kappa fn/3}$. The probability of failure of the third assertion is bounded by $e^{-l \hat{f}n/3} + \delta f$.

Using union bound the total probability of success is at least:
$$1-\kappa - \delta f - e^{-\kappa fn/3} - e^{-lf_kn/3}-e^{-l \hat{f}n/3}$$
Using $f_k,\hat{f}\geq f$ and factorizing:
$$1-\kappa - \delta f - e^{-fn/3}\left(e^\kappa+2e^l\right)$$
If we want to use the original parameters $c$ and $\beta$:
\begin{align*}
1- &\left(4e^{-(c^2/4)}/c^2\right) -\delta f\\
- &e^{-fn/3}\left(e^{4e^{-(c^2/4)}/c^2}+2e^{e^{-\beta}/(1-4e^{-(c^2/4)}/c^2)}\right)    
\end{align*}

We will later use values of $c=3,\beta=4$, which would give probability bounds of:
$$1-0.0468-\delta f - 3.09e^{-fn/3}$$
Notice how using reasonable values $f=0.01, n=2000, \delta=0.1$ we get a bound of probability $94.8\%$.

\underline{Bounding the distance}
We know both $p$ and $o$ belong to the shell. Moreover both have lowerbounds on their density; which can only lower their distance to each other (and to the center of the Gaussian). Their distance is thus upperbounded by our result in lemma \ref{appendix:lemma:kappa} changing the denominator from $(1-\kappa)$ to $(1-\kappa)^2$ because now both $p$ and $o$ are restricted to be in the shell:
$$P\left(dist(p,o)\geq U\right) \leq \frac{e^{-\beta}}{(1-\kappa)^2}$$
Adding this to the total bound we get that with probability at least $1-\kappa - \delta f - \frac{e^{-\beta}}{(1-\kappa)^2} - e^{-fn/3}\left(e^\kappa+e^l\right)$ the distance from $p$ to the closest output divided by $r_f(p)$ is:
$$\frac{dist(p,o)}{r_f(p)} \leq \frac{U}{L-2c} \leq \frac{\sqrt{2} \sqrt{d+2\sqrt{d\beta}+2\beta}}{\sqrt{2} \sqrt{d-2\sqrt{d\beta}}-2c}$$
Note that the series expansion as $d\rightarrow\infty$ converges to 1:
$$\frac{dist(p,o)}{r_f(p)} \leq 1 + c\sqrt{\frac{2}{d}} + O(d^{-1})$$

\underline{Bounding the radius of the output}
We know that $r_f(p)\geq L-2c$ and $r_f(o) \leq U+2c$. Therefore:
$$\frac{r_f(o)}{r_f(p)}\leq \frac{U+2c}{L-2c} \leq \frac{\sqrt{2} \sqrt{d+2\sqrt{d\beta}+2\beta}+2c}{\sqrt{2} \sqrt{d-2\sqrt{d\beta}}-2c}$$
Note that the series expansion as $d\rightarrow\infty$ converges to 1:
$$\frac{r_f(o)}{r_f(p)} \leq 1 + 2c\sqrt{\frac{2}{d}} + O(d^{-1})$$

\textbf{Second case}: the probability of a point $p$ s.t. $r_{(1-\epsilon)(1-2l)f}(p)> r$ but $p$ has an output nearby is at most $\kappa+e^{-fn/3}$



Let us upperbound the proability of \HAC\ giving an output in a Gaussian of $f_k\geq (1-\epsilon)f$ simply by 1. With probability $(1-\kappa)$ point $p$ is in the shell of $Z_k$ and with probability at least $1-e^{-fn/3}$ the empirical quantile $1-2l$ of $dist(p,q)$ is at most $r$.

Joining both probabilities by union bound we get that the probability of a point $p$ satisfying both $p\in \text{shell}$ and quantile $\phi_{1-2l}\leq r$ is at least $\left(1-\kappa-e^{-fn/3}\right)$. 

If the point is in the shell and its empirical quantile $1-2l$ is less than $r$ but $r_{(1-\epsilon)(1-2l)f}(p)>r$, that means: $$(1-\epsilon)(1-2l)f>(1-2l)f_k \Rightarrow f_k < (1-\epsilon)f$$
This implies that $p$ is in a Gaussian with mass less than $f$ and therefore no output $o$ will be nearby.
\QEDA

\begin{conjecture}
We conjecture that $\Phi_x(r\in \text{shell}) \approx \Phi_x(\sqrt{d-1}+(2x-1)c)$.
This allows us to improve the guarantees of theorem \ref{appendix:thm:high-dim-inf-dist} by lowering $\Delta_1$ from\\ $\frac{\sqrt{2} \sqrt{d+2\sqrt{d\beta}+2\beta}}{\sqrt{2} \sqrt{d-2\sqrt{d\beta}}-2c}-1=c\sqrt{\frac{2}{d}} + O(d^{-1})$ to $\frac{\sqrt{2} \sqrt{d+2\sqrt{d\beta}+2\beta}}{\sqrt{2} \sqrt{d-2\sqrt{d\beta}}-2lc}-1 = lc\sqrt{\frac{2}{d}}+O(d^-1)$ and $\Delta_2$ from\\ $\frac{\sqrt{2} \sqrt{d+2\sqrt{d\beta}+2\beta}+2c}{\sqrt{2} \sqrt{d-2\sqrt{d\beta}}-2c} -1=2c\sqrt{\frac{2}{d}} + O(d^{-1}) $ to $\frac{\sqrt{2} \sqrt{d+2\sqrt{d\beta}+2\beta}+2lc}{\sqrt{2} \sqrt{d-2\sqrt{d\beta}}-2lc} -1=2lc\sqrt{\frac{2}{d}}+O(d^{-1})$.
\label{appendix:conjecture:totally_non_concave}
\end{conjecture}

Increasing $r$ shifts the whole distribution, increasing all quantiles. This implies that lower $r$ will have more impact on lower quantiles and bigger $r$ will have more impact on bigger quantiles. In particular we can make the extreme approximation of $\Phi_x(r)$ being a delta function with all its mass at one point, independendent of $q$. This would make all its quantiles equal and also $\Phi_x(r\in \text{shell}) = \Phi_x(\sqrt{d-1}+(2x-1)c)$. Although this extreme approximation is unlikely to be true, it may give a better estimate than not knowing where the intersection is at all.

Now the small quantile $l$ is close to $\sqrt{d-1}-c$, in particular at $\sqrt{d-1}-c+2lc$ and the bigger quantile is close to $\sqrt{d-1}+c$, at $\sqrt{d-1}+c-2lc$. This allows us to substitute the factors $2c$ by $2lc$ since now the 1-Lipschitz cone starts within $2lc$ of the edge of the shell. Thus:
$$\Delta_1 = \frac{\sqrt{2} \sqrt{d+2\sqrt{d\beta}+2\beta}}{\sqrt{2} \sqrt{d-2\sqrt{d\beta}}-2lc}-1 = lc\sqrt{\frac{2}{d}}+O(d^{-1})$$
$$\Delta_2 = \frac{\sqrt{2} \sqrt{d+2\sqrt{d\beta}+2\beta}+2lc}{\sqrt{2} \sqrt{d-2\sqrt{d\beta}}-2lc} -1=2lc\sqrt{\frac{2}{d}}+O(d^{-1})$$

\begin{conjecture}
Theorem \ref{appendix:thm:high-dim-inf-dist} still holds if Gaussian means are at distance at least $\Omega(d^{1/4})$ instead of infinitely far away.
\end{conjecture}
As derived in the same online draft\footnote{\url{https://www.cs.cmu.edu/~venkatg/teaching/CStheory-infoage/chap1-high-dim-space.pdf}} of \cite{blum2016foundations} as lemma \ref{appendix:lemma:kappa}, two Gaussians of variance 1 can be separated if they are at least $d^{1/4}$ apart because most pairs of points in the same Gaussian are at distance $\sqrt{2d}+O(1)$ and most pairs of points in different Gaussians are at distance $\sqrt{|\mu_1-\mu_2|^2+2d}+O(1)$. For the lowerbound on inter-Gaussian distance to be bigger than the upperbound on intra-Gaussian distance we need:

\begin{align*}
\sqrt{2d}+O(1) &\leq \sqrt{|\mu_1-\mu_2|^2+2d}+O(1) \\
\Rightarrow 2d + O(d^{1/2}) &\leq 2d + |\mu_1-\mu_2|^2 \\
\Rightarrow |\mu_1-\mu_2| &\in \Omega(d^{1/4}) \text{is enough}
\end{align*}

If that is the case the typical intra-Gaussian distance is $\sqrt{2d}$ and the typical inter-Gaussian distance is $\sqrt{2d+\Omega(\sqrt{d})}$.

We modify their calculations a bit to get concrete numbers. In particular we approximate $|O(1)|\approx 3\sqrt{2}$ since we used $c=3$ in our realistic bounds (see \ref{appendix:subsec:realistic_bounds}). This gives us:
$$\sqrt{2d}+3\sqrt{2} \leq \sqrt{|\mu_1-\mu_2|^2+2d}-3\sqrt{2}$$
$$|\mu_1-\mu_2|\geq 2\sqrt{6}\sqrt{\sqrt{d}+3}$$

\begin{figure}
    \centering
    \includegraphics[width=0.4\textwidth]{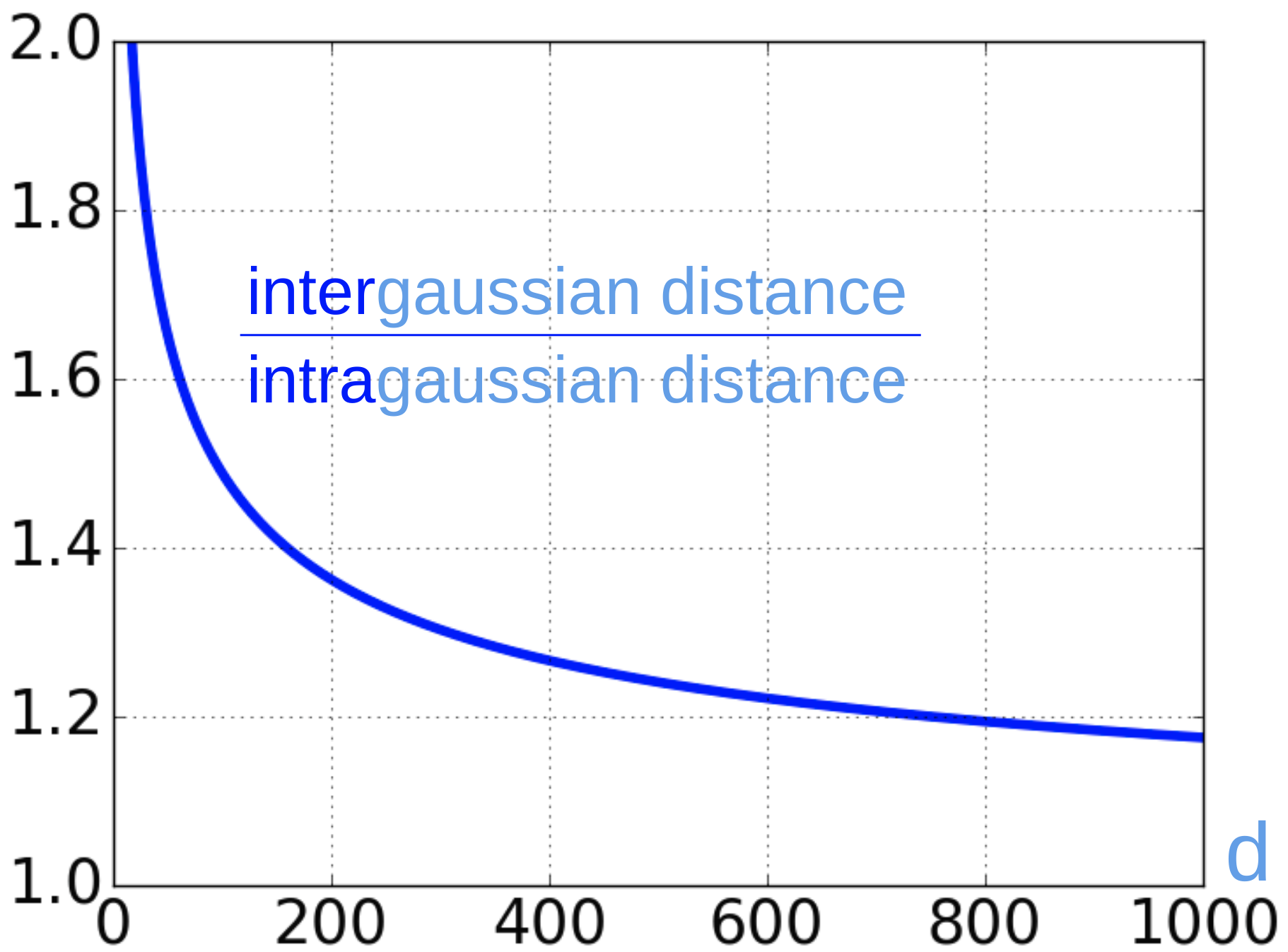}
    \caption[Ratio between distances of different clusters and same clusters]{Ratio between distances of different clusters and same clusters. These distances can be quite similar in high dimensions (only 30\% bigger) for our guarantees to still be valid. In particular note that in our experiments in section \ref{sec:people-id} the distance ratio is $0.65/0.5=1.3$.}
    \label{appendix:fig:inter_intra_Gaussian_distances}
\end{figure}

Now, again the typical intra-Gaussian distance is $\sqrt{2d}$ and the typical inter-Gaussian distance is $\sqrt{2d+24\sqrt{d}+3}$. Their ratio can be seen in figure \ref{appendix:fig:inter_intra_Gaussian_distances}.

We conjecture that if we can separate two clusters the amount of points of other clusters within distance $r_f(p)$ will be exponentially small and thus, in essence, is as if they were infinitely far away; which would make our original theorem \ref{appendix:thm:high-dim-inf-dist} applicable.

\paragraph*{Inserting realistic numbers in our bounds}
\label{appendix:subsec:realistic_bounds}

\begin{figure}
    \centering
    \begin{subfigure}[t]{.8\linewidth}
    \begin{center}
        \includegraphics[width=\linewidth]{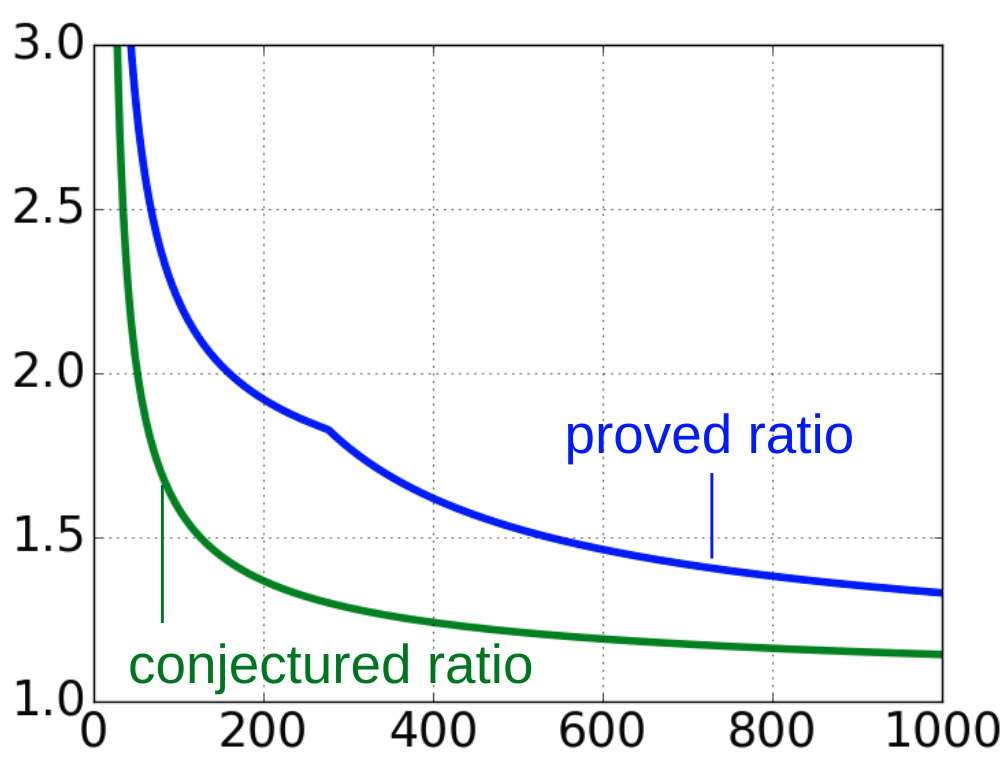}
    \end{center}
    \caption[width=\linewidth]{$dist(p,o)/r_f(p)$ guarantees}
    \end{subfigure}
    \hfill
    \begin{subfigure}[t]{.8\linewidth}
    \begin{center}
        \includegraphics[width=\linewidth]{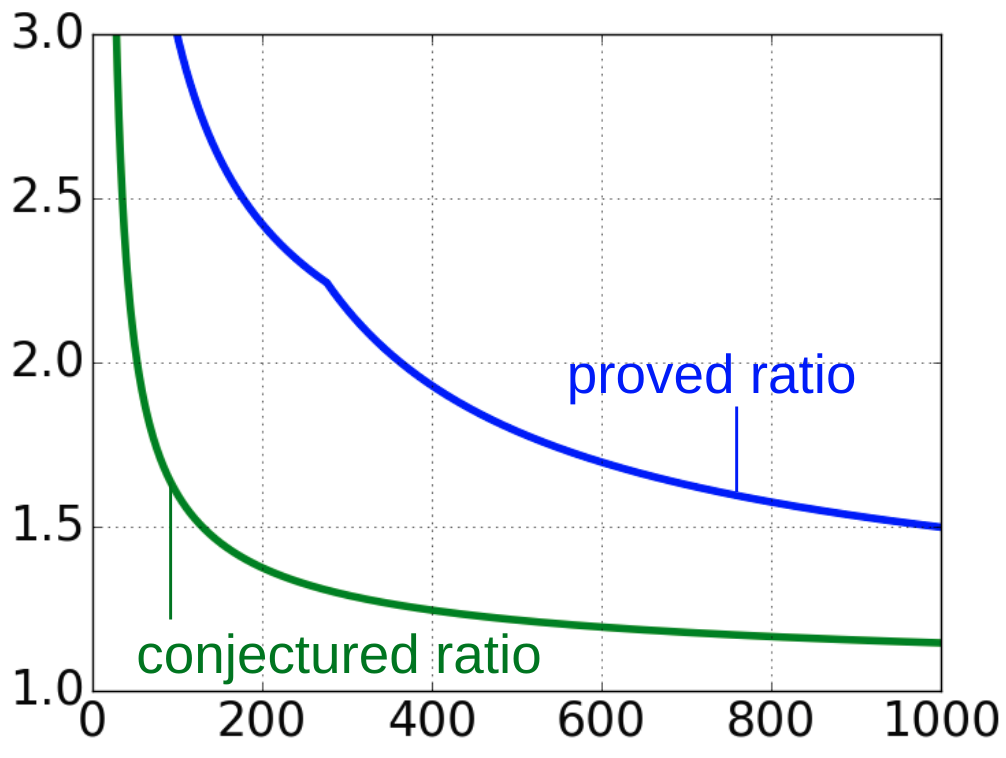}
    \end{center}
    \caption[width=\linewidth]{$r_f(o)/r_f(p)$ guarantees}
    \end{subfigure}
    \caption{Guarantees for high dimensions with realistic parameters.}
    \label{appendix:fig:plot_guarantees_ratios}
\end{figure}

First, we note an optimization to guarantees that we didn't include in the theorem since it didn't have an impact on asymptotic guarantees. In theorem \ref{appendix:thm:high-dim-inf-dist} we set $L=\sqrt{2}\sqrt{d-4\sqrt{d}}-2c$ which we proved using 1-Lipschitzness; however, we can also argue that $\Phi_x(r)$ is monotonically increasing and therefore $L$ should be lowerbounded by $\Phi_x(0)= \sqrt{d-4\sqrt{d}}$ which is easy to compute since it depends on the \textit{central} chi distribution. Therefore, in practice, the denominators in the guarantees are: $$\max\left(\sqrt{2}\sqrt{d-4\sqrt{d}}-6, \sqrt{d-4\sqrt{d}}\right)$$
This max produces the kinks in the blue lines in figure \ref{appendix:fig:plot_guarantees_ratios}.

In the same setting as before, let $c=3, \beta=4, \delta=0.1, n=2000, f=0.01$; which are typical values we could use in experiments. We have $\kappa = 4/c^2 e^{-c^2/4} < 0.047$, $L = \sqrt{2}\sqrt{d-4\sqrt{d}, }U = \sqrt{2}\sqrt{d+4\sqrt{d}+8}$, $l=e^{-\beta}/(1-\kappa) < 0.0193$.
Plug in values in the bounds on the theorem above we get:

Any point $p$ with $r_{1.05f}(p)\leq r$ will have an output within distance $\frac{\sqrt{2} \sqrt{d+4\sqrt{d}+8}}{\max\left(\sqrt{2}\sqrt{d-4\sqrt{d}}-6, \sqrt{d-4\sqrt{d}}\right)}r_f(p)$ and radius at most $\frac{\sqrt{2} \sqrt{d+4\sqrt{d}+8}+6}{\max\left(\sqrt{2}\sqrt{d-4\sqrt{d}}-6, \sqrt{d-4\sqrt{d}}\right)}r_f(p)$ with probability at least $94.6\%$. Using conjecture \ref{appendix:conjecture:totally_non_concave} we predict an output within distance $\frac{\sqrt{2} \sqrt{d+4\sqrt{d}+8}}{\sqrt{2}\sqrt{d-4\sqrt{d}}-0.12}r_f(p)$ and radius at most $\frac{\sqrt{2} \sqrt{d+4\sqrt{d}+8}+0.12}{\sqrt{2} \sqrt{d-4\sqrt{d}}-0.12}r_f(p)$. These guarantees are plotted as a function of $d$ in figure \ref{appendix:fig:plot_guarantees_ratios}.

The probability of a point $p$ satisfying both $r_{0.96(1-\epsilon)f}>r$ and having an output within distance $r$ is at most: $e^{-4}/(1-0.047)+e^{-20/3}<0.021$.
\subsection{Time scaling}
We have described a time-independent version of the algorithm, where all points regardless of when they came, have equal weight. However, it is easy to extend this algorithm to make point $i$ have weight proportional to $e^{-\frac{t-t_i}{\kappa}}$ for any timescale $\tau$, where $t$ is the current time $t$ and $t_i$ is the time when point $i$ was inserted. We will see our algorithm requires $t_i\geq t_{i-1}$, inputs coming in non-decreasing times, a very natural constraint.

By construction, the last point inserted will still have weight $1$. Now, let $t'$ be the time of the last inserted point. We can update all the weights of the already received points multiplying by $e^{-\frac{t-t'}{\tau}}$. Therefore all weights can be updated by the same multiplication. Since everyone is multiplied by the same number, sums of weights can also be updated by multiplying by $e^{-\frac{t-t'}{\tau}}$.

We now only need to worry about jumps. We can keep a counter for the total amount of weight of points for the points received until now. Let us call $w_{p_j, t_k}$ to the weight of point $p_j$ at the time point $k$ arrives. Since we want to have a uniform distribution over those weights, when the $i$-th point arrives we simply assign the probability of jumping to $\frac{1}{\sum_{j\leq i} w_{p_j, t_i}}$. Note that for the previous case of all weights $1$ (which is also the case of $\tau = \infty$) this reduces to the base case of probability $\frac{1}{i}$.

We have checked the last point has the correct probability, what about all the others? Let us pick point $j < i$, its probability is:

\begin{align*}
& \frac{1}{\sum_{k \leq j} w_{p_k, t_j}} \cdot \left(1 - \frac{1}{\sum_{k\leq j+1} w_{p_k, t_j}}\right)\cdots \left(1 - \frac{1}{\sum_{k\leq i} w_{p_k, t_i}}\right) \\
&= \frac{1}{\sum_{k \leq j} w_{p_k, t_j}} \cdot \left(\frac{\sum_{k \leq j} w_{p_k, t_{j+1}}}{\sum_{k\leq j+1} w_{p_k, t_{j+1}}}\right)\cdots \left(\frac{\sum_{k\leq i-1} w_{p_k, t_{i}}}{\sum_{k\leq i} w_{p_k, t_i}}\right) \\
 &= \frac{1}{\sum_{k \leq j} w_{p_k, t_j}} \cdot \left(\frac{e^{-\frac{t_{j+1}-t_j}{\tau}}\sum_{k \leq j} w_{p_k, t_{j}}}{\sum_{k\leq j+1} w_{p_k, t_j}}\right)\cdots \\
 &\cdots \left(\frac{e^{-\frac{t_i-t_{i-1}}{\tau}}\sum_{k\leq i-1} w_{p_k, t_{i-1}}}{\sum_{k\leq i} w_{p_k, t_i}}\right)
\end{align*}
This is a telescoping series which becomes:
\begin{align*}
&1\cdot e^{-\frac{t_{j+1}-t_j}{\tau}} \cdot e^{-\frac{t_{j+2}-t_{j+1}}{\tau}} \cdots e^{-\frac{t_{i}-t_{i-1}}{\tau}}\cdot  \frac{1}{\sum_{k\leq i} w_{p_k, t_i}}\\
& = \frac{e^{-\sum_{k={j+1}}^i\frac{t_k-t_{k-1}}{\tau}}}{\sum_{k\leq i} w_{p_k, t_i}} = \frac{e^{-\frac{t_i-t_j}{\tau}}}{\sum_{k\leq i} w_{p_k, t_i}}
\end{align*}
Note that the numerator is the exact weight point $p_j$ should have at time $t_i$ and thus all points have their probabilities of being an output point proportional to their weights. Moreover, note that each multiplying factor, which is the probability of not hopping at every added point, must be between 0 and 1. This forces the condition $t_{j+1}\geq t_j \forall j$; in other words, we must feed the observations in non-decreasing order, a very natural condition.

Finally, all our proofs use general assertions about weights and probabilities, without assuming those came from discrete elements. Thus, we can use \textit{fraction of weights} instead of \textit{fraction of points} in all the proofs and they will all still hold.

\subsection{Fixing the number of outputs}
We currently have two ways of querying the system: 1) Fix a single distance $r$ and a frequency threshold $f$, and get back all regions that are $(r,f)$-dense; 2) Fix a frequency $f$, and return a set of points $\{p_i\}$, each with a different radius $\{r_i\}$ s.t. a point $p$ near output point $p_i$ is guaranteed to have $r_f(p) \approx r_i$.

It is sometimes more convenient to directly fix the number of outputs instead. With \HAC\ we go one step further and return a list of outputs sorted according to density (so, if you want $o$ outputs, you pick the first $o$ elements from the output list). Here are two ways of doing this: 1) Fix radius $r$. Find a set of outputs $p_i$ each $(r,f_i)$-dense. Sort $\{p_i\}$ by decreasing $f_i$, thus returning the densest regions first. For example, in our people-finding experiment (section \ref{sec:people-id}) we know two points likely correspond to the same person if their distance is below 0.5. We thus set $r=0.5$ and sort the output points by their frequencies using that radius, thus getting a list of characters from \textit{most} to \textit{least} popular.
2) Fix frequency $f$, sort the list of regions from smallest to biggest $r$. Note, however, that the algorithm is given a fixed memory size which governs the size of the possible outputs and the frequency guarantees.

In general, it is useful and easy to remove duplicates with this framework. Moreover, we can do so keeping our guarantees with minimal changes. 
\begin{theorem}
\label{appendix:thm:awesome-postprocess}
We can apply a post-processing algorithm that takes parameter $\gamma$ in time $\Theta\left(\frac{\log (f\delta)}{\epsilon f^2}\right)$ to reduce the number of output points to $(1+2\epsilon)/f$ while guaranteeing that for any point $p$ there is an output within $(4\gamma + 3)r_f(p)$. For $\gamma=1.25$ this guarantees within $8r_f(p)$. 
The same algorithm guarantees that for any $(r_{max}, f)$-\textit{dense} point there will be an output within $7r_{max}$.
\end{theorem}
Let us sort the set of outputs $O$ in any order. Then, for any output $o\in O$ we add it to the filtered list of outputs $O^*$ if and only if its $B(o,2r)\cap B(o^*,2r) = \emptyset \forall o^*\in O^*$. By construction, we have a list of balls that do not intersect and each has at least $(1-\epsilon)f$ fraction of points. The fraction contained in the union of those balls is at most 1 and they do not intersect, thus the number of balls is at most $\frac{1}{(1-\epsilon)f} \leq \frac{1+2\epsilon}{f}$. From here we can see that iterating for every output and comparing it to any point in the list is $\Theta\left(\frac{\log (f\delta)}{\epsilon f}\right)\Theta\left(\frac{1+2\epsilon}{f}\right) = \Theta\left(\frac{\log (f\delta)}{\epsilon f^2}\right)$.

Now, for any \textit{dense} point $p$, we know:
\begin{itemize}
    \item $\exists p^* \in D^*$ s.t. $d(p,p^*) \leq 2r$
    \item $\exists o\in O$ s.t. $d(p^*,o) \leq r$
    \item $\exists o^* \in O^*$ s.t. $d(o,o^*) \leq 4r$
\end{itemize}
Adding all those distances and applying triangular inequality we know that for any dense point $p$ $\exists o$ s.t. $d(p,o^*)\leq 7r$.
\QEDA

\begin{theorem}
We can reduce the number of output points to $\frac{1+2\epsilon}{f}$ such that any point has an output within $(4\gamma+3)r_f$ For $\gamma = 1.25$ this guarantees within $8r_f(p)$.
\label{appendix:thm:awesome-postprocess-multiple-radii}
\end{theorem}
We will follow an argument similar to theorem \ref{appendix:thm:awesome-postprocess}; however it will be slightly trickier because we have multiple radii.

Again, we know that with probability at least $1-\delta$ any point $p$ has an output within distance $3r_f(p)$ of radius at most $2\gamma r_f(p)$. Let us assume we're in this situation and show how we can apply a postprocessing to filter the points. We denote the output radius of an output $o$ by $rad(o)$

We sort all outputs by their output radius in increasing order and breaking ties arbitrarily. We iterate through this ordered set of outputs $O$. For any output $o$ we add it to a filtered output $O^*$ if and only if $B(o, rad(o))\cap B(o^*,rad(o^*)) = \emptyset \forall o^*\in O$. By definition the balls of points in $O^*$ do not intersect and each contains at least $(1-\epsilon)f$ fraction of points. Therefore the fraction of points contained in the union is the sum of the fractions and this fraction must be at most 1. Therefore there are at most $\frac{1}{(1-\epsilon)f} \leq \frac{1+2\epsilon}{f}$ filtered outputs.

Now, for any point $p$, we know $\exists o\in O\text{ s.t. }d(p,o)\leq 3r_f(p)$ and $rad(o)\leq 2\gamma r_f(p)$. Then,
If $o\in O^*$ then we have shown $\exists o^*\in O^*$ s.t. $d(p,o)\leq 3r_f(p)\leq (4\gamma+3)r_f(p)$ and with radius at most $2\gamma r_f(p)$.

Otherwise, $o \not \in O^*$. Then, by construction, its ball intersects with some $o^*\in O^*$ with $rad(o^*)\leq rad(o)$. Therefore:
$$\exists o^*\in O^*\text{ s.t. }rad(o^*)\leq rad(o) \leq 2\gamma r_f(p)$$ and 
$$d(o,o^*)\leq 2\cdot 2\gamma r_f(p) \Rightarrow$$ $$d(p,o^*) \leq d(p,o)+d(o,o^*)\leq (4\gamma+3)r_f(p)$$

\QEDA

Notice that the number of outputs is arbitrarily close to the optimal $\frac{1}{f}$.

\end{document}